\documentclass[12pt]{article}
\usepackage{arxiv}
\usepackage{graphicx} 
\usepackage[T1]{fontenc}
\usepackage{amsmath}
\usepackage[english]{babel}
\usepackage{array,tabularx}
\usepackage{booktabs} 
\usepackage{makeidx}  
\usepackage{xparse}
\usepackage{moredefs}
\usepackage{lips}
\usepackage{epstopdf} 
\usepackage{color}
\usepackage{csquotes}
\usepackage{xcolor}
\usepackage{times}
\usepackage[hidelinks]{hyperref}
\usepackage[nolist,nohyperlinks]{acronym}
\usepackage{comment}
\usepackage{paralist}
\usepackage{cleveref}
\usepackage{wrapfig}
\usepackage{multirow}
\usepackage{microtype}
\usepackage{listings}
\usepackage[super]{nth}
\usepackage[numbers]{natbib}

\usepackage{float}

\usepackage[style=base,labelfont=bf,labelsep=period,tableposition=top,font=small]{caption}
\expandafter\def\expandafter\UrlBreaks\expandafter{\UrlBreaks
  \do\a\do\b\do\c\do\d\do\e\do\f\do\g\do\h\do\i\do\j%
  \do\k\do\l\do\m\do\n\do\o\do\p\do\q\do\r\do\s\do\t%
  \do\u\do\v\do\w\do\x\do\y\do\z\do\A\do\B\do\C\do\D%
  \do\E\do\F\do\G\do\H\do\I\do\J\do\K\do\L\do\M\do\N%
  \do\O\do\P\do\Q\do\R\do\S\do\T\do\U\do\V\do\W\do\X%
  \do\Y\do\Z}

\newcolumntype{L}[1]{>{\raggedright\arraybackslash}p{#1}}   
\newcolumntype{C}[1]{>{\centering\arraybackslash}p{#1}}     
\newcolumntype{R}[1]{>{\raggedleft\arraybackslash}p{#1}}    

\title{TS-Arena - A Live Forecast Pre-Registration Platform}

\newif\ifuniqueAffiliation

\ifuniqueAffiliation 
\author{ \href{https://orcid.org/0009-0005-9136-8525}{\includegraphics[scale=0.06]{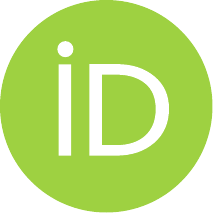}
\hspace{1mm}Marcel Meyer}\\
	\And
	\href{https://orcid.org/0000-0002-8411-6347}{\includegraphics[scale=0.06]{orcid.pdf}\hspace{1mm}Sascha Kaltenpoth} \\
    \And
    \href{https://orcid.org/0009-0004-8388-619X}{\includegraphics[scale=0.06]{orcid.pdf}\hspace{1mm}Henrik Albers} \\    
    \And
	\href{https://orcid.org/0009-0000-8515-5711}{\includegraphics[scale=0.06]{orcid.pdf}\hspace{1mm}Kevin Zalipski} \\

    \And
	\href{https://orcid.org/0000-0002-0369-1607}{\includegraphics[scale=0.06]{orcid.pdf}\hspace{1mm}Oliver Müller} \\
}
\else
\usepackage{authblk}

\newbox{\orcid}\sbox{\orcid}{\includegraphics[scale=0.06]{orcid.pdf}} 
\author[1]{%
	\href{https://orcid.org/0009-0005-9136-8525}{\usebox{\orcid}\hspace{1mm}Marcel Meyer\thanks{\texttt{marcel.meyer@uni-paderborn.de}}}%
}
\author[1]{%
	\href{https://orcid.org/0000-0002-8411-6347}{\usebox{\orcid}\hspace{1mm}Sascha Kaltenpoth}%
}
\author[1]{%
	\href{https://orcid.org/0009-0004-8388-619X}{\usebox{\orcid}\hspace{1mm}Henrik Albers}%
}
\author[1]{%
	\href{https://orcid.org/0009-0000-8515-5711}{\usebox{\orcid}\hspace{1mm}Kevin Zalipski}%
}
\author[1]{%
	\href{https://orcid.org/0000-0002-0369-1607}{\usebox{\orcid}\hspace{1mm}Oliver Müller}%
}
\affil[1]{Paderborn University, Data Analytics Group}
\fi

\renewcommand{\shorttitle}{\textit{arXiv} Template}

\begin{document}   

\maketitle
\setcounter{footnote}{0}

\begin{abstract} 
Time Series Foundation Models (TSFMs) are transforming the field of forecasting. However, evaluating them on historical data is increasingly difficult due to the risks of train-test sample overlaps and temporal overlaps between correlated train and test time series. To address this, we introduce TS-Arena, a live forecasting platform that shifts evaluation from the known past to the unknown future. Building on the concept of continuous benchmarking, TS-Arena evaluates models on future data. Crucially, we introduce a strict forecasting pre-registration protocol: models must submit predictions before the ground-truth data physically exists. This makes test-set contamination impossible by design. The platform relies on a modular microservice architecture that harmonizes and structures data from different sources and orchestrates containerized model submissions. By enforcing a strict pre-registration protocol on live data streams, TS-Arena prevents information leakage and offers a faster alternative to traditional static, infrequently repeated competitions (e.g. the M-Competitions). First empirical results derived from operating TS-Arena over one year of energy time series demonstrate that established TSFMs accumulate robust longitudinal scores over time, while the continuous nature of the benchmark simultaneously allows newcomers to demonstrate immediate competitiveness. TS-Arena provides the necessary infrastructure to assess the true generalization capabilities of modern forecasting models. The platform and corresponding code are available at \url{https://ts-arena.live/}.
\end{abstract}
\section{Introduction}
Time series forecasting is vital in domains such as finance \cite{zhang_multi-period_2025}, energy \cite{meyer_benchmarking_2025}, and operations \cite{klee2025measuring}. Forecasting is currently undergoing a paradigm shift towards Time Series Foundation Models (TSFMs): large neural network models trained on broad, heterogeneous collections of time series that can be directly applied in a zero-shot manner \cite{liang_foundation_2024}, following the idea of scaling laws as observed in large language models (LLMs) \cite{kaplan_scaling_2020,yao2025towards}. These models are able to produce forecasts for new or unseen series without task-specific training or fine-tuning and have demonstrated zero-shot performance across domains that is superior or comparable to that of models requiring retraining \cite{aksu_gift-eval_2024,li_tsfm-bench_2025}.

\begin{figure*}[b!]
    \centering
    \includegraphics[width=0.7\linewidth]{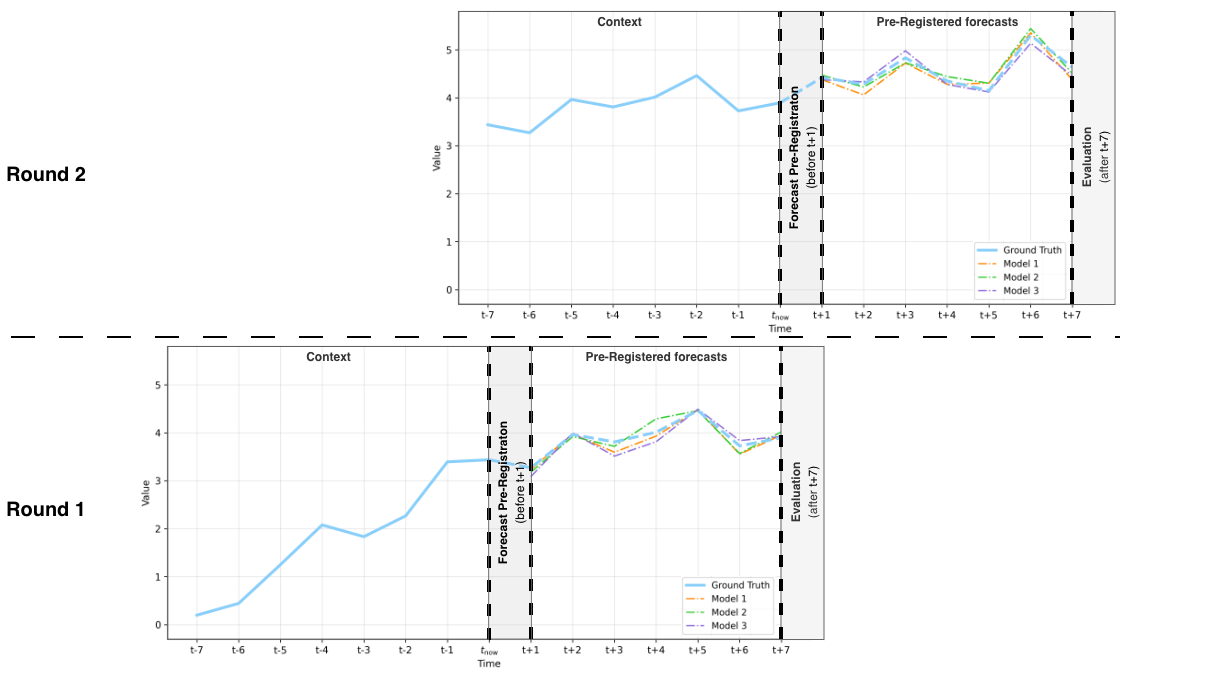}
    \caption{The Forecast Pre-Registration Protocol}
    \label{fig:fprp}
\end{figure*}

However, the paradigm of foundation models introduces two novel evaluation challenges related to information leakage \cite{aksu_gift-eval_2024, meyer_time_2025,rodrigo_data_2024}. Direct information leakage, also referred to as test data contamination, was already thoroughly researched in LLMs \cite{liao_rethinking_2023}. Training LLMs on large opaque datasets caused accidential inclusion of benchmark test data in the LLMs' pre-training data, leading to the LLM ``evaluation crisis'' \cite{liao_rethinking_2023}. This can also occur when training samples of one model overlap with test samples used for another model, invalidating cross-model comparisons, a phenomenon frequently observed in TSFMs \cite{aksu_gift-eval_2024, meyer_time_2025}. In contrast, indirect temporal information leakage is inherent to the temporal characteristics of forecasting. It arises when TSFMs are evaluated on held-out test series whose forecast horizons \emph{temporally} overlap with training series from \emph{correlated} time series, allowing implicit access to future information through shared temporal structure \cite{meyer_time_2025,rodrigo_data_2024}. An example for this are highly correlated bus and metro usage counts from the same period \cite{rodrigo_data_2024}. If the bus time series is in the TSFMs training data and the metro time series is in a benchmark's test data, a pre-trained model can capture future patterns from the bus usage and transfer it to the metro usage forecasts. In a real-world forecasting scenario a model naturally does not have access to this temporally leaked information. Such temporal leakage effect is even present in less correlated time series \cite{meyer_time_2025}. Recent studies have shown that both types of information leakage can substantially inflate reported TSFM performance, leading to unreliable conclusions about model generalization \cite{aksu_gift-eval_2024,meyer_time_2025,rodrigo_data_2024}.

As outlined in recently published benchmarking requirements \cite{meyer_benchmarking_2025}, established evaluation methods such as time-based splits, held-out datasets, or competitions are prone to these special forms of information leakage or impractical due to their long-term timeline \cite{bojer_kaggle_2021,makridakis_m6_2024}.
Considering this, recent benchmarks made valuable contributions to TSFM evaluation in general, by substantially expanding model coverage, dataset diversity, and evaluation regimes \cite{aksu_gift-eval_2024,li_tsfm-bench_2025}, but did not or only partially address information leakage concerns. They enabled a holistic view of forecasting performance by evaluating heterogeneous model families---including foundation models, deep learning methods, and domain-specific forecasters---under zero-shot, few-shot, and full-shot settings, and across diverse domains, frequencies, forecast horizons, and data characteristics such as trend, seasonality, and stationarity \cite{aksu_gift-eval_2024, li_tsfm-bench_2025}. In terms of leakage prevention, some benchmarks employ fixed train–test splits or held-out evaluation datasets intended to be excluded from future model training \cite{aksu_gift-eval_2024}, while others introduce entirely new evaluation datasets or API-restricted datasets that cannot accidentally be incorporated into the training data \cite{goktas2025tempusbench,xu_fidel-ts_2025}.

Yet, existing approaches cannot fully address information leakage concerns while at the same time allowing the required flexibility for training TSFMs: Introducing fixed pre-training and test splits \cite{aksu_gift-eval_2024} inhibits the flexibility for pre-training TSFMs with broad and heterogeneous time series---a core factor of the TSFM scaling laws \cite{yao2025towards}. Furthermore, existing benchmarks all rely on historical data and retrospective evaluation \cite{aksu_gift-eval_2024, li_tsfm-bench_2025, goktas2025tempusbench, xu_fidel-ts_2025}. In particular, they remain vulnerable to direct or indirect information leakage when published test samples are inadvertently used to train TSFMs, or when models are trained on correlated time series from the same time period as the benchmark test sets, such as bus and metro usage case \cite{rodrigo_data_2024}, or a more subtile example of weather data in training data and solar generation in test data from the same region and time period. Thus, leveraging any fixed dataset that is not evolving over time and directed into the future---regardless of how carefully curated---can eventually lead to information leakage \cite{liang_foundation_2024,yao2025towards}. We are not aware of any benchmark mechanism which addresses directly the time series forecasting specific forms of information leakage.

To address the inherent challenges of information leakage and test-set contamination, we propose a framework that leverages the unique generative nature of time series data: the fact that observations are continuously produced in real-time. We develop from this our core principle drawn from competitions as benchmarks \cite{bojer_kaggle_2021,makridakis_m6_2024}: a forecast pre-registration protocol (FPRP) that enforces the submission of forecasts before the real-world ground-truth data exists, forecasting into the real, unknown future. By evaluating the pre-registered forecasts immediatley after the ground-truth is known, this protocol makes both direct and indirect information leakage impossible by design.

\section{Concept: Forecast Pre-Registration Protocol}
\label{pre-registration-mechanism}
Similar to the concept of \textit{living} benchmarks \cite{erickson_tabarena_2025}, TS-Arena adapts the core idea of continuous maintenance and evolving leaderboards to the specific characteristics of time series forecasting. Here, \textit{live} implies not merely a ranking that is continuously updated with new models and datasets, but a dynamic test environment where the evaluation targets are generated in real-time by the unfolding future.

While current \textit{living} benchmarks rely on retrospective train-test splits of static, typically historical, datasets, we propose a novel forecasting pre-registration protocol (FPRP). It uses the evaluation time $t$ as a strict temporal split point, denoted as $t_{now}$, that advances continuously in real-time. Forecasting models have full access to historical ground-truth data up to $t_{now}$ and must commit their forecasts for fixed future horizon $H$ before the first future ground-truth value of the target time series becomes available. This protocol is continuously applied in rounds, which refer to small fast-paced competitions \cite{bojer_kaggle_2021,makridakis_m6_2024}. The protocol for a single forecast round proceeds as follows:

\begin{enumerate}
    \item \textbf{Round Initialization:} At a predefined point in time, a new round is initialized. A round consists of a set of one or more time series with common characteristics. Each round has a context of previous observations of a time series, a registration window, and a forecast horizon.
    \item \textbf{Context Acquisition:} The latest context of previous observations for each time series is provided to round participants. It contains the historical ground truth of context [$X_{t_{now}-C}, \dots, X_{t_{now}}$] with length $C$ available up to the current moment.
    \item \textbf{Inference and Registration:} Participants generate forecasts for the defined horizon [$t_{now}+1, \dots, t_{now}+H$]. These predictions must be uploaded on the platform within the registration window.
    \item \textbf{Window Closure:} The registration window is determined by the duration from the occurrence of one observation to another (i.e., the frequency, e.g., 15 minutes or 1 hour) and ranges maximum from [$t_{now}, \dots, t_{now}+1$). It closes before the next ground-truth value $X_{t_{now}+1}$ is observed and recorded by the system. Any submission attempted after window closure is strictly rejected.
    \item \textbf{Evaluation:} Once the future ground-truth values [$X_{t_{now}+1},$ $ \cdots, X_{t_{now}+H}$] for the full horizon have materialized, the system computes the final error metrics.
\end{enumerate}

While a round's forecast horizon's $H$ actual values are sequentially observed and recorded in real time, a follow-up round is initialized with a registration later than $t_{now}$. This process repeats iteratively and theoretically infinitely. Figure \ref{fig:fprp} illustrates our forecast pre-registration protocol in two rounds using a context length of seven time steps and a horizon of seven time steps as an example. In our example we shift by the horizon $H$ of the first round, but the registration start of a new round can be arbitrary after $t_{now}$.

By strictly closing the submission window before the next future data point of the time series $X_{t_{now}+1}$ comes into existence, the test set remains logically inaccessible during the inference process. This prevents any form of information leakage and look-ahead bias by design, while the iterative repetition or rounds enables a fast and accurate evaluation in sense of time series cross-validation \cite{hyndman_forecasting_2021}.

\section{Implementation: The TS-Arena Microservice Architecture}
\subsection{Overall Architecture}
\begin{figure*}[h!]
    \centering
    \includegraphics[width=0.8\linewidth]{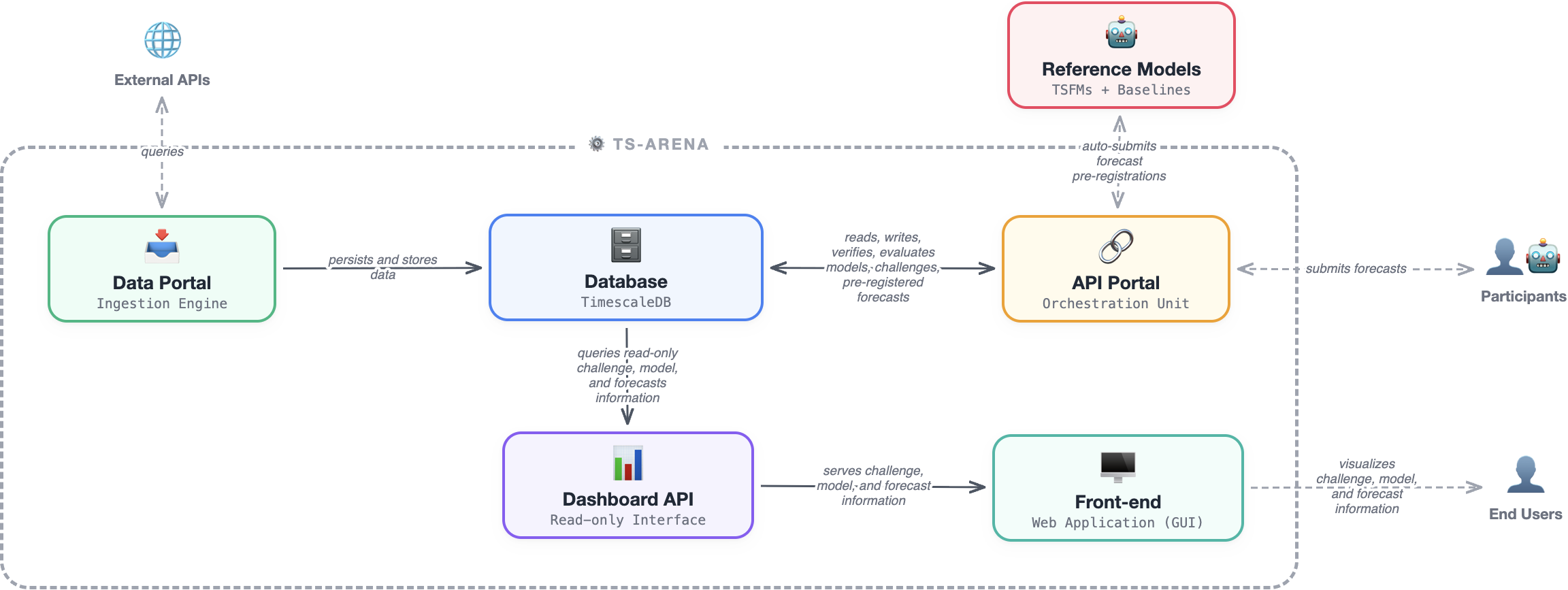}
    \caption{The Microservice Architecture}
    \label{fig:ts-arena}
\end{figure*}

To implement the FPRP, TS-Arena draws on a modular microservice architecture using a Docker Compose network, illustrated in Figure \ref{fig:ts-arena}. We further bundle multiple time series in challenges for specific characteristics, such as same domains, frequencies, context lengths, or horizons. Within a challenge round, the FPRP is implemented so that every challenge has iteratively starting rounds.

While the microservice architecture ensures scalability, dependency isolation, and clear separation between data ingestion, management, and evaluation, the concept of challenges, including the FPRP, enables information leakage free evaluation of TSFMs. The primary services of the architecture are defined as follows:

\begin{description}
    \item[Database:] The database, a TimescaleDB\footnote{https://github.com/timescale/timescaledb} instance, handles the storage and retrieval of high-frequency time series data. It provides a structured storage for incoming live data, forecast submissions, and user metadata.
    
    \item[Data Portal:] This service acts as decoupled ingestion engine. It queries external APIs for continuous retrieval of time series, normalizes heterogeneous data formats into the internal schema, and converts timestamps to UTC while at the same time persisting the original timezone information as an extra attribute. It is responsible for correctly storing the raw as well as the processed data in the database, following the Slowly Changing Dimension Type 2 (SCD2) archiving strategy for data warehouses \cite{kimball_data_2013}.
    
    \item[API Portal:] The API Portal serves as the central orchestration unit and interface for participants. It manages user authentication, creates challenge rounds, validates incoming forecast submissions against active registration windows, and triggers evaluation routines as soon as new ground truth data becomes available.
    
    \item[Dashboard API:] The dedicated Dashboard (read-only) API supplies the front-end with challenge statuses, live time series data, submitted forecasts, and continuously updated leaderboard information.

    \item[Front-end:]
    The front-end provides a graphical user interface (GUI) for end users in form of a web app that provides visualizations of challenges, challenge and model information, and leaderboards to track live rounds.
    
    \item[Reference Model Service:] To provide a neutral evaluation context for model performance, this service hosts containerized implementations of state-of-the-art TSFMs and statistical baselines. These models act as neutral participants, autonomously requesting context from the API Portal and submitting forecasts to it. This guarantees that every challenge round contains a robust set of forecasts for immediate comparison.
\end{description}

\subsection{Database and Data Portal}
While our reliance on real-time data presents greater acquisition effort compared to static historical datasets, the resulting timeliness is essential for realistic forecasting.
In principle, any time series with (near) real-time accessibility can be integrated into TS-Arena. For the sake of clarity, the platform's first long-term challenges focus on time series from the energy domain. The energy domain provides an ideal starting point, combining high-frequency data availability \cite{kanter_gridstatus_2025} with complex and dynamic time series patterns \cite{hong_energy_2020,meyer_benchmarking_2025}, thereby challenging the forecasting capabilities of TSFMs. Furthermore, energy time series originate from diverse data-generating processes, include physical processes like in wind and solar power generation, human behavior reflected in load and consumption, and economic market mechanisms in the case of energy prices \cite{hong_energy_2020,lago_forecasting_2021,meyer_benchmarking_2025}. Based on these characteristics we built challenges for load, generation, and price forecasting for different frequencies and forecast horizons. The challenges further include data from different aggregation levels, ranging from market bidding zone over transmission system operator to trading hub level. The data for these challenges is sourced and aggregated from international transmission and independent system operators. We query them from providers such as SMARD, Fingrid and Gridstatus, which includes New York and California ISO. We refer to the providers in the following. The challenges are summarized in Table \ref{tab:challenge_schedules_refined}.
All time series data is refined and standardized into a unified schema. We consistently ingest data at the highest source granularity (e.g., 3-minute intervals). To ensure consistency across different time frames, we leverage TimescaleDB continuous aggregates to sample the data down into uniform frequencies, such as 15-minute or 1-hour blocks. A critical component of our data archival is the use of a SCD2 archiving strategy \cite{kimball_data_2013}. This allows the system to reconstruct the exact information state available at any historical $t_{now}$. Consequently, researchers can verify what data was actually available to a model at the specific moment of inference, distinct from subsequently corrected data, ensuring full reproducibility of the live conditions.

\begin{table*}[t!]
    \centering
    \small
    \addtolength{\tabcolsep}{-3pt} 
    \begin{tabular}{@{}l l c c l l r@{}}
        \toprule
        \textbf{Description} & \textbf{Category} & \textbf{\# TS} & \textbf{Ctx} & \textbf{Freq} & \textbf{Horiz} & \textbf{Registration (UTC)} \\
        \midrule
        SMARD Day-Ahead Price & price & 15 & 1000 & 15m & 1d & 02:00 pm - 02:14 pm \\
        SMARD Net Load & consum. & 10 & 1000 & 15m & 1d & 08:00 am - 08:14 am\\
        SMARD Generation & generat. & 24 & 1000 & 15m & 1d & 08:00 pm - 08:14 pm\\
        SMARD Day-Ahead Price & price & 15 & 1000 & 1h & 3d & 03:00 pm - 03:59 pm\\
        SMARD Net Load & consum. & 10 & 1000 & 1h & 3d & 09:00 am - 09:59 am\\
        SMARD Generation & generat. & 24 & 1000 & 1h & 3d & 09:00 pm - 09:59 pm\\
        \addlinespace
        Gridstatus Hub Price & price & 18 & 1000 & 15m & 1d & 04:00 pm - 04:14 pm \\
        Gridstatus Load & consum. & 11 & 1000 & 15m & 1d & 09:00 am - 09:14 am \\
        Gridstatus Generation & generat. & 17 & 1000 & 15m & 1d & 08:00 pm - 08:14 pm \\
        Gridstatus Hub Price & price & 18 & 1000 & 1h & 3d & 05:00 pm - 05:59 pm\\
        Gridstatus Load & consum. & 11 & 1000 & 1h & 3d & 10:00 am - 10:59 am \\
        Gridstatus Generation & generat. & 17 & 1000 & 1h & 3d & 09:00 pm - 09:59 pm \\
        \addlinespace
        FINGRID Challenge & mixed & 12 & 1000 & 15m & 1d & 04:45 am - 04:59 am\\
        FINGRID Challenge & mixed & 12 & 1000 & 1h & 3d & 05:00 am - 05:59 am\\
        \bottomrule
    \end{tabular}
    \caption{Overview of Challenge Schedules (All domains: Energy). Ctx = Context.}
    \label{tab:challenge_schedules_refined}
\end{table*}

\subsection{API Portal}
\subsubsection{Orchestration by Challenges}
As mentioned, we adapt the concept of competitions \cite{bojer_kaggle_2021,makridakis_m6_2024} and convert it into more fast-paced \textit{challenges}. A challenge bundles multiple time series and contains many rounds, whose forecasts need to be pre-registered before the same time point $t_{now}$ and evaluated in an aggregated manner after the actual target values exist. 

The time series of a challenge usually exhibit comparable characteristics. At least the frequency and the forecast horizon are consistent within a challenge to facilitate inference. Further criteria can be added such as aggregation level, (sub-)category (e.g., load, generation), unit of measurement, or any other available metadata to create domain-, frequency-, or task-specific challenges (e.g., long-term forecasting or household electricity forecasts).

Following our FPRP, the challenges are conducted in \textit{rounds}. A round represents a single execution of the FPRP, as described in Section \ref{pre-registration-mechanism}. The rounds of a challenge are initialized at pre-defined and publicly visible points in time each day, following a specific challenge schedule. The currently scheduled challenges, summarized in Table \ref{tab:challenge_schedules_refined}, are continuously hosted by the API Portal.

\subsubsection{Evaluation Protocol and Aggregation}
For each forecast we calculate the Mean Absolute Scaled Error ($MASE$) as a scale independent metric with the naive forecast as the baseline \cite{hyndman_another_2006}. Similar to TabArena \cite{erickson_tabarena_2025}, we report an $ELO$ score with confidence intervals, inspired by chess ratings, through a pairwise comparison of the models based on their $MASE$ scores per challenge round. Like a chess grandmaster's rating versus a beginner's rating, the confidence intervals show how reliable performances are. These confidence intervals are calculated via bootstrapping, as explained in the subsequent section \ref{sec:uncertainty}. Established models, which participated in many challenge rounds, have narrow intervals, while newer models, which submitted less forecasts, have wide intervals, warning against treating short-term success as proven superiority. This is based on our calculation strategy: let $\mathcal{R}$ be the set of discrete evaluation rounds (e.g., daily submission cycles). For each challenge round $r \in \mathcal{R}$ and model $m_i$, we compute $P_{i,r}$, which denotes to the mean MASE across the series in that round.

\subsubsection{Intersection-Based Pairwise Updates}
We address the asynchronous entry of challenge round participation by implementing a strict pairwise model validity check. For any two models $m_i$ and $m_j$, the pairwise comparison in round $r$ is conducted if both models participated in that round ($P_{i,r}$ and $P_{j,r}$ are valid).
The outcome $S_{i,j,r}$ of such a pairwise comparison is defined as:

\begin{equation}
S_{i,j,r} = 
\begin{cases} 
1.0 & \text{if } P_{i,r} < P_{j,r} \quad (\text{Win}) \\
0.0 & \text{if } P_{i,r} > P_{j,r} \quad (\text{Loss})
\end{cases}
\end{equation}

It is extremely unlikely that two models will have exactly the same MASE, but we still handle this case in the code with a draw. Consequently, established models continue to refine their ratings based on their full history, while comparisons with new entrants are effectively restricted to the overlapping evaluation windows without discarding global historical data.

\subsubsection{Uncertainty Quantification via Randomized Replay}
\label{sec:uncertainty}
Since the $ELO$ rating is an online learning algorithm, the final scores can exhibit path dependency, that is, sensitivity to the temporal order of match outcomes. To quantify epistemic uncertainty and strictly test the stability of the rankings, we employ a \textit{Randomized Replay Bootstrap} ($B=500$) similar to TabArena \cite{erickson_tabarena_2025}. Instead of sampling datasets, we permute the chronological order of the evaluation rounds $\mathcal{R}$ across $B$ iterations. For each iteration, the entire history is replayed in a randomized sequence, recalculating the $ELO$ trajectory from scratch. We calculate a 95\% confidence interval by deriving the 2.5th and 97.5th percentiles from these bootstrap iterations. Models with a long, consistent history ($|\mathcal{T}_i| \gg 0$) will converge to a stable rating regardless of the sequence order, resulting in narrow confidence intervals. Conversely, recent entrants with few forecasts will exhibit higher volatility under permutation, rightfully yielding wider confidence intervals that reflect their provisional status. The $ELO$ is reported by challenges, frequencies, and forecast horizons. Additionally, we report the aggregated mean $MASE$ together with its standard deviation.

While the current implementation focuses on predicting point estimates, the architecture supports the future integration of probabilistic scoring rules (e.g., $CRPS$) to evaluate uncertainty quantification.

\subsection{Dashboard API \& Front-End}
\begin{figure*}[h!]
    \centering
    \includegraphics[width=0.75\linewidth]{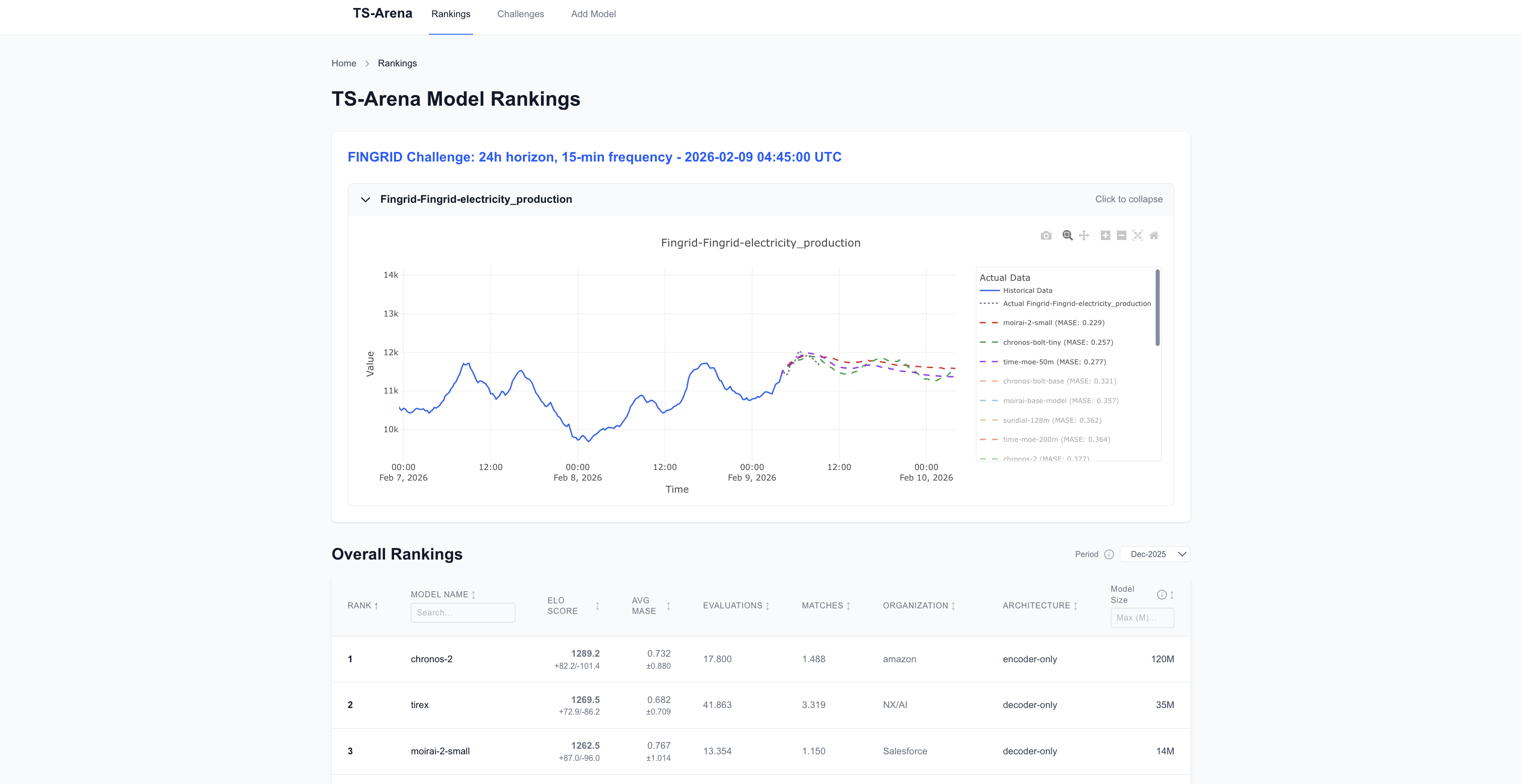}
    \caption{TS-Arena Challenge and Model View}
    \label{fig:ts-arena-front-end}
\end{figure*}

The Dashboard API provides detailed information about every time series and every challenge but also aggregated information. It is used by the front-end shown in Figure \ref{fig:ts-arena-front-end}. The front-end comprises an aggregated leaderboard and forecast views for currently running challenges as well as specific views for challenges and models. The challenges can be filtered according to diverse characteristics, enabling researchers and practitioners to identify models of interest for their specific field of research or practice.
\subsection{Reference Model Service}
The Reference Model Service ensures an active participation of a number of reference models in every challenge and round that serve as common baselines. The Reference Model Service interacts with the public TS-Arena API Portal just like any other participant, ensuring the reference models operate under the exact same constraints (e.g., submission windows, data access) as other (external) participants.

We adhere to a strict ``Standard Implementation'' protocol to guarantee reproducibility and fair comparison:
\begin{itemize}
    \item \textbf{Official Codebases:} We utilize the official repositories provided by the model authors.
    \item \textbf{Zero-Shot Configurations:} TSFMs are executed using author-recommended default settings to evaluate their out-of-the-box generalization capabilities without domain-specific tuning.
\end{itemize}

To establish a robust and representative leaderboard, we initialize our benchmark with a diverse selection of TSFMs alongside fundamental statistical baselines. Our selection criteria for TSFMs prioritizes models described in peer-reviewed literature with downloadable model checkpoints and good performance in other benchmarking studies. We also include successor and recent pre-publication models. In addition, we strive for architectural diversity to rigorously assess how performance differs in different zero-shot forecasting scenarios.

\begin{table}[h]
\centering
\small
\begin{tabular}{l l c l l}
\hline
\textbf{Model} & \textbf{Architecture} & \textbf{Review} & \textbf{Published} & \textbf{Source} \\
\hline
TTMs$^{*^1}$ R1 
  & Mixed
  & 1 
  & 08.01.24 
  & \cite{ekambaram_tiny_2024} \\

Moirai 
  & Encoder-only
  & 1 
  & 04.02.24 
  & \cite{woo_unified_2024} \\

Time-MoE 
  &  Decoder-only
  & 1 
  & 24.09.24 
  & \cite{shi_time-moe_2024} \\

TTMs$^{*^1}$ R2 
  & Mixed
  & 1 
  & 08.10.24 
  & \cite{ekambaram_tiny_2024} \\

Chronos Bolt 
  & Encoder-decoder
  & 0 
  & 26.11.24 
  & \cite{ansari2024fast} \\

TimesFM 2.0 
  & Decoder-only 
  & 0 
  & 20.12.24 
  & \cite{das_decoder-only_2024} \\

Sundial 
  &  Decoder-only
  & 1 
  & 02.02.25 
  & \cite{liu_sundial_2025} \\

TiRex 
  &  Encoder-only
  & 1 
  & 29.05.25 
  & \cite{auer_tirex_2025} \\

VisionTS++ 
  & Vision
  & 0 
  & 06.08.25 
  & \cite{shen_visionts_2025} \\

Flowstate 
  & Encoder-decoder
  & 0 
  & 07.08.25 
  & \cite{graf_flowstate_2025} \\

TimesFM 2.5 
  & Decoder-only 
  & 0 
  & 15.09.25 
  & \cite{das_decoder-only_2024} \\

Chronos-2 
  & Encoder-only 
  & 0 
  & 17.10.25 
  & \cite{ansari_chronos-2_2025} \\

Moirai 2 
  & Encoder-only
  & 0 
  & 12.11.25 
  & \cite{liu_moirai_2025} \\
\hline
\end{tabular}
\caption{Reference TSFMs included in TS-Arena}
{$^{*^1}$TTMs = TinyTimeMixers}
\label{tab:ts_foundation_models}
\end{table}

Our initial set includes decoder-only models (TimesFM 2.0 and 2.5, Time-MoE, Moirai 2, Sundial), encoder-only models (Chronos 2, Moirai and Moment), encoder-decoder models (Chronos Bolt, Flowstate), xLSTM-based models like TiRex, MLP-based models like TinyTimeMixers, and the Vision Transformer model VisionTS++. Table \ref{tab:ts_foundation_models} summarizes the models, their architectures, their current peer review status, and their publishing date. In our initial TSFM set, we excluded models which are not out-of-the-box suitable for zero-shot forecasting and must be fine-tuned, like the Moment family \cite{goswami_moment_2024}.

To ensure a comprehensive comparative analysis, we benchmark the foundation models against fundamental statistical baselines, such as the naive and seasonal naive heuristics. By juxtaposing TSFMs against these simple statistical baselines, we aim to provide both the research community and practitioners with a grounded perspective on where TSFM zero-shot capabilities genuinely outperform simple baselines in live settings.

\section{Results and Analysis}
\subsection{Initial Setup}
In our initial setup, we start with 186 live time series from three different data sources: SMARD \cite{bundesnetzagentur_smard_nodate}, gridstatus \cite{kanter_gridstatus_2025} and FINGRID \cite{oyj_fingrid_nodate}. We organized these time series into 14 challenges. For each challenge, a new round is started at a different time each day, so that users have regular opportunities to participate, while also distributing the compute load more evenly throughout the day. To populate the platform with initial data, we backtested the year 2025 on the TS-Arena platform with these challenges and with the references models. A full year covers all seasons and seasonalities for quarter-hourly and hourly data. In the spirit of our leakage-free future data evaluation approach, the hosted reference TSFMs started participating in challenges only after their release date. As an example, Chronos-2 was published on October 17th, 2025 \cite{ansari_chronos-2_2025}. Therefore, it retrospectively started participating in challenges starting from that date, regardless its training data cut-off date. This approach led to to over 5,000 challenge rounds in 2025, totaling in 3.928.191 test set points. This is, of course, only a current snapshot that automatically evolves over time. To support transparency, independent verification, and faster offline experimentation, we additionally release an archive (context, ground truth and forecasts) as a public dataset on HuggingFace at \url{https://huggingface.co/datasets/DAG-UPB/TS-Arena-Archive}. The archive is refreshed each quarter so that researchers can always evaluate against the most recent snapshot, including newly released TSFMs. The first release covers data starting Q1 2026. Further details and the rules of use are provided in Appendix \ref{appendix:archive}.

Data regarding electricity generation from hard coal and natural gas are currently excluded from this benchmark. Due to high marginal costs and CO$_2$ pricing, these energy sources typically operate at the margin of the merit order rather than providing baseload power and are more suitable to cover the residual load during periods of low renewable generation \cite{bertsch_flexibility_2016}. Consequently, the resulting time series show extended periods of constant (zero) generation, with an occasional high-magnitude spike strongly dependent on exogenous factors, such as weather conditions and current power demand. Illustrations of these irregular patterns are provided in \autoref{sec:ts_selection}. Since there are currently no covariates provided in any of the live data APIs, the scope of this benchmark is currently limited to univariate time series forecasting. Thus, the coal and natural gas data were omitted to ensure a fair evaluation of the models and their predictions. We plan to re-introduce these series once the benchmark supports covariates.

\vspace{2pt}

\begin{table*}[h]
\centering
\small
\begin{tabular}{llcccccc}
\toprule
& & \multicolumn{2}{c}{\textbf{Global}} & \multicolumn{2}{c}{\textbf{15 min / 1 day}} & \multicolumn{2}{c}{\textbf{1 h / 3 days}} \\
\cmidrule(lr){3-4} \cmidrule(lr){5-6} \cmidrule(lr){7-8}
Model & Rounds & ELO$^{(\text{CI})}$ & MASE$_{(\pm\text{std})}$ & ELO$^{(\text{CI})}$ & MASE$_{(\pm\text{std})}$ & ELO$^{(\text{CI})}$ & MASE$_{(\pm\text{std})}$ \\
\midrule
chronos-2 & 1488 & $\mathbf{1289^{+82}_{-101}}$ & $0.732_{{\pm 0.88}}$ & $\mathbf{1288^{+82}_{-92}}$ & $0.706_{{\pm 1.17}}$ & $\mathbf{1295^{+90}_{-99}}$ & $0.751_{{\pm 0.58}}$ \\
tirex & 3319 & $1270^{+73}_{-86}$ & $\mathbf{0.682}_{{\pm 0.71}}$ & $1277^{+76}_{-72}$ & $\mathbf{0.685}_{{\pm 0.89}}$ & $1273^{+86}_{-92}$ & $\mathbf{0.680}_{{\pm 0.48}}$ \\
moirai-2-small & 1150 & $1263^{+87}_{-96}$ & $0.767_{{\pm 1.01}}$ & $1261^{+89}_{-97}$ & $0.734_{{\pm 1.46}}$ & $1270^{+78}_{-92}$ & $0.788_{{\pm 0.58}}$ \\
timesfm-2.5-200m & 1902 & $1240^{+86}_{-102}$ & $0.747_{{\pm 0.92}}$ & $1282^{+79}_{-87}$ & $0.721_{{\pm 1.20}}$ & $1216^{+88}_{-96}$ & $0.768_{{\pm 0.61}}$ \\
chronos-bolt-base & 5090 & $1214^{+94}_{-88}$ & $0.723_{{\pm 0.98}}$ & $1262^{+70}_{-95}$ & $0.727_{{\pm 1.19}}$ & $1180^{+97}_{-87}$ & $0.718_{{\pm 0.72}}$ \\
flowstate & 2411 & $1214^{+89}_{-102}$ & $0.710_{{\pm 0.92}}$ & $1241^{+93}_{-103}$ & $0.696_{{\pm 1.20}}$ & $1203^{+98}_{-96}$ & $0.722_{{\pm 0.58}}$ \\
chronos-bolt-small & 5090 & $1206^{+86}_{-97}$ & $0.726_{{\pm 0.97}}$ & $1233^{+86}_{-86}$ & $0.739_{{\pm 1.20}}$ & $1183^{+87}_{-89}$ & $0.713_{{\pm 0.67}}$ \\
chronos-bolt-mini & 5090 & $1200^{+82}_{-106}$ & $0.724_{{\pm 0.92}}$ & $1212^{+86}_{-92}$ & $0.742_{{\pm 1.12}}$ & $1187^{+71}_{-86}$ & $0.705_{{\pm 0.67}}$ \\
chronos-bolt-tiny & 5090 & $1194^{+74}_{-91}$ & $0.722_{{\pm 0.89}}$ & $1197^{+96}_{-86}$ & $0.746_{{\pm 1.10}}$ & $1192^{+78}_{-98}$ & $0.699_{{\pm 0.60}}$ \\
timesfm-2.0-500m & 5088 & $1178^{+95}_{-89}$ & $0.737_{{\pm 1.00}}$ & $1168^{+86}_{-82}$ & $0.779_{{\pm 1.31}}$ & $1197^{+78}_{-98}$ & $0.694_{{\pm 0.54}}$ \\
sundial-128m & 4828 & $1154^{+90}_{-80}$ & $0.736_{{\pm 0.90}}$ & $1195^{+78}_{-72}$ & $0.753_{{\pm 1.14}}$ & $1118^{+83}_{-85}$ & $0.720_{{\pm 0.57}}$ \\
moirai-base-model & 5090 & $1113^{+73}_{-84}$ & $0.753_{{\pm 0.56}}$ & $1066^{+96}_{-93}$ & $0.796_{{\pm 0.61}}$ & $1162^{+79}_{-87}$ & $0.709_{{\pm 0.51}}$ \\
moirai-large & 5090 & $1099^{+88}_{-92}$ & $0.774_{{\pm 0.72}}$ & $1021^{+107}_{-89}$ & $0.838_{{\pm 0.88}}$ & $1167^{+75}_{-91}$ & $0.710_{{\pm 0.52}}$ \\
tinytimemixer-r2-1024-96 & 4890 & $1046^{+99}_{-77}$ & $0.790_{{\pm 1.08}}$ & $1078^{+89}_{-84}$ & $0.819_{{\pm 1.35}}$ & $1014^{+99}_{-85}$ & $0.760_{{\pm 0.70}}$ \\
tinytimemixer-r1-1024-96 & 5088 & $1042^{+87}_{-87}$ & $0.785_{{\pm 1.00}}$ & $1053^{+92}_{-78}$ & $0.817_{{\pm 1.26}}$ & $1034^{+95}_{-89}$ & $0.753_{{\pm 0.64}}$ \\
time-moe-50m & 5089 & $1035^{+97}_{-90}$ & $0.823_{{\pm 1.67}}$ & $1011^{+81}_{-79}$ & $0.896_{{\pm 2.25}}$ & $1058^{+100}_{-97}$ & $0.749_{{\pm 0.72}}$ \\
time-moe-200m & 5088 & $1033^{+96}_{-85}$ & $0.824_{{\pm 1.62}}$ & $1003^{+80}_{-83}$ & $0.896_{{\pm 2.16}}$ & $1063^{+93}_{-89}$ & $0.752_{{\pm 0.75}}$ \\
moirai-small & 5090 & $1031^{+98}_{-83}$ & $0.788_{{\pm 0.61}}$ & $1005^{+93}_{-96}$ & $0.837_{{\pm 0.71}}$ & $1054^{+94}_{-84}$ & $0.740_{{\pm 0.47}}$ \\
seasonal-naive & 5105 & $922^{+117}_{-112}$ & $0.975_{{\pm 1.30}}$ & $913^{+118}_{-108}$ & $1.046_{{\pm 1.70}}$ & $924^{+128}_{-126}$ & $0.903_{{\pm 0.70}}$ \\
seasonal-average & 5101 & $866^{+114}_{-108}$ & $1.086_{{\pm 2.56}}$ & $822^{+119}_{-100}$ & $1.245_{{\pm 3.49}}$ & $903^{+112}_{-110}$ & $0.927_{{\pm 0.93}}$ \\
visiontspp-base & 2422 & $862^{+103}_{-84}$ & $1.026_{{\pm 1.27}}$ & $893^{+89}_{-84}$ & $0.978_{{\pm 1.66}}$ & $820^{+111}_{-83}$ & $1.068_{{\pm 0.77}}$ \\
visiontspp-large & 2422 & $797^{+99}_{-79}$ & $1.101_{{\pm 1.58}}$ & $825^{+91}_{-78}$ & $1.096_{{\pm 2.10}}$ & $765^{+99}_{-77}$ & $1.105_{{\pm 0.93}}$ \\
simple-moving-average & 5089 & $609^{+96}_{-74}$ & $1.431_{{\pm 4.96}}$ & $594^{+101}_{-73}$ & $1.606_{{\pm 6.88}}$ & $610^{+103}_{-83}$ & $1.256_{{\pm 1.32}}$ \\
\bottomrule
\end{tabular}
\caption{Reference TSFM and Statistical Baseline MASEs and Elo-Scores in the Backtesting Year 2025}
\label{tab:elo_global}
\end{table*}

\subsection{Overall model performance}
Table \ref{tab:elo_global} presents the global $ELO$ ratings and $MASE$ by the end of the year 2025 for the evaluated TSFMs and baseline methods. The table distinguishes between overall performance and performance at specific frequencies and horizons.

The top positions on the leaderboard are all occupied by TSFMs. \texttt{chronos-2} ranks first with a global ELO of $1289^{+82}_{-101}$, followed by \texttt{tirex} ($1270^{+73}_{-86}$) and \texttt{moirai-2-small} ($1263^{+87}_{-96}$). The confidence intervals (indicated by subscripts/superscripts) vary across models due to the number of participated rounds. \texttt{tirex} has completed 3319 rounds, resulting in narrower bounds ($+73/-86$) compared to \texttt{chronos-2}, which has completed 1488 rounds ($+82/-101$). In general, the confidence intervals of the top models overlap, indicating a similar performance and the need for more evaluations.

The highest-ranking baseline method is \texttt{seasonal-}\textit{naive} with an ELO of $922$. Simple heuristic methods such as the \texttt{simple-}\textit{moving-}\textit{average} are positioned at the bottom of the leaderboard (ELO $609$).

The model rankings differ slightly between the two evaluated resolutions. In the high-frequency setting (15 min / 1 day), \texttt{chronos-2} holds the highest rating ($1288$), followed closely by \texttt{timesfm-}\textit{2.5-}\textit{200m} ($1282$) and \texttt{tirex} ($1277$). Similarly, in the lower-frequency setting (1 h / 3 days), \texttt{chronos-2} holds the highest rating ($1295$), but this time followed by \texttt{tirex} ($1273$) and \texttt{moirai-2-small} ($1270$).

As we included different model sizes of each model family, we are able to analyze the impact of parameter scaling on zero-shot forecasting performance. For example, within the \texttt{chronos-}\textit{bolt} family, ELO ratings monotonically increase with model size: \texttt{tiny} ($1194$) $<$ \texttt{mini} ($1200$) $<$ \texttt{small} ($1206$) $<$ \texttt{base} ($1214$).

Newer iterations of model families rank higher than their predecessors. \texttt{moirai-2-small} ($1263$) is rated higher than \texttt{moirai-}\textit{base-}\textit{model} ($1113$) and \texttt{moirai-large} ($1099$). Similarly, \texttt{timesfm-}\textit{2.5-}\textit{200m} ($1240$) is rated above \texttt{timesfm-2.0-500m} ($1178$). The ELO rankings generally align with the MASE scores, though differences exist. \texttt{tirex} achieves the lowest global MASE ($0.682$) despite ranking second in ELO. Judging from the confidence intervals of the MASE, \texttt{tirex} exhibits less overall deviation from the ground truth, whereas larger models achieve stronger fits on some target series but weaker fits on others, resulting in more wins and higher ELO rankings. \texttt{sundial-128m} shows a competitive performance with an ELO of $1154$ and a MASE of $0.736$, placing it above \texttt{moirai-}\textit{base-}\textit{model} ($0.753$ MASE, $1113$ ELO).

Our results indicate that TSFMs yield the lowest average performance in energy generation, while achieving their best results in energy consumption. More detailed results and analysis by challenges can be found in the Appendix Section \ref{appendix:results}.

\section{Related Work}
Recent benchmarks made valuable contributions toward valid and reliable evaluation of TSFMs. In particular, large-scale benchmarking efforts like TSFM-Bench expand model coverage to include LLM-based TSFMs, time-series–pretrained foundation models, and domain-specific forecasting models, and evaluate them under zero-shot, few-shot, and full-shot regimes \cite{li_tsfm-bench_2025}. TSFM-Bench further diversifies evaluation data across domains such as electricity, environment, and banking, as well as across statistical characteristics including trend, seasonality, and stationarity \cite{li_tsfm-bench_2025}, providing a holistic and representative view of forecasting performance. Yet, TSFM-Bench does neither address nor discuss potential information leakage. Their evaluation dataset comprises well-known datasets that are commonly used for pre-training and evaluation of TSFMs, potentially leading to direct information leakage through an overlap of pre-training data of new TSFMs with the test data from TSFM-Bench \cite{goktas2025tempusbench, meyer_benchmarking_2025}.

GIFT-EVAL emphasizes diversity along complementary axes, including data frequency, forecast horizon, prediction length, and model families spanning statistical, deep learning, and foundation models \cite{aksu_gift-eval_2024}. Furthermore, GIFT-EVAL introduces a fixed test set intended to be excluded from future TSFM training to prevent information leakage. As stated earlier, the fixed pre-training set leads to a lack of flexibility for scaling TSFMs with new data \cite{yao2025towards}, while the historical test data split can also become part of a new TSFMs pre-training dataset, leading to information leakage. Although GIFT-EVAL features a flag for potential information leakage, this does not facilitate meaningful model comparison, as their own experiments show that information leakage can inflate reported performance metrics \cite{aksu_gift-eval_2024}. However, without a leakage-free TSFM, the actual effect size of information leakage remains unknown in their leaderboard.

To explicitly tackle information leakage, TempusBench utilizes less commonly used datasets to mitigate contamination of test sets \cite{goktas2025tempusbench}. In a similar way, Fidel-TS addresses the challenge by leveraging authentication-protected APIs to periodically gather new evaluation data \cite{xu_fidel-ts_2025}. Nevertheless, these approaches do not rule out that theses datasets become part of future TSFMs pre-training data and, consequently, compromise benchmark validity.

\newpage
\section{Discussion \& Limitations}
Current benchmarks represent valuable progress in TSFM evaluation practice, and substantially improve upon earlier forecasting benchmarks by broadening model coverage, introducing standardized protocols, and enabling systematic comparison across heterogeneous methods \cite{aksu_gift-eval_2024,goktas2025tempusbench,li_tsfm-bench_2025,xu_fidel-ts_2025}.

Still, existing benchmarking approaches are prone to classical test set contamination but also to direct and indirect information leakage. In particular, sample overlaps between train and test series make valid cross-model evaluation impossible, and temporal overlaps between train and correlated test series can inflate model performance by leaking future information \cite{meyer_benchmarking_2025, rodrigo_data_2024}. 

Our study addresses these concerns by providing the forecast pre-registration protocol (FPRP) enabled by the usage of live and future data, thereby preventing information leakage due to the physical nonexistence of ground-truth test data.

Nevertheless, the current scope of TS-Arena is intentionally narrower in terms of available datasets and configuration space. While our 3.9 million data points from 2025 clearly outnumber the 86.858 data points of TSFM-Bench \cite{li_tsfm-bench_2025}, they are relatively small compared to the over 300 million data points in Gift-Eval \cite{aksu_gift-eval_2024}. While this might be seen as a temporary limitation of our approach, TS-Arena is inherently extensible. As a living benchmark, it continuously incorporates new data sources, frequencies, horizons, and challenge types as new real-world data is generated. In this sense, TS-Arena complements existing benchmarks: while static benchmarks provide breadth and controlled analysis, TS-Arena provides a forward-only, leakage-free evaluation infrastructure that always remains its integrity as models and data evolve.

Furthermore, a deeper look into the dynamics of the live leaderboard validates the ``living benchmark'' paradigm. In the simulated year 2025, we observe on our TS-Arena platform that newly introduced models can immediately achieve top positions, yet their wide confidence intervals indicate a lack of statistical certainty. This highlights that a sustained evaluation period is necessary to distinguish between genuine generalization capability and temporary over-performance due to specific market regimes.

\section{Conclusion \& Outlook}
We presented TS-Arena, a live benchmark platform for Time Series Foundation Models (TSFMs) that enforces a Forecast Pre-Registration Protocol (FPRP): Forecasts must be submitted before the corresponding ground truth exists. This simple temporal commitment eliminates benchmark leakage by design and enables continuous, transparent model comparison on evolving real-world data. This approach is currently the only that prevents information leakage in TSFMs.

While our living benchmark already comprises 186 different energy time series evaluated in over 5000 challenge rounds, leading to over 3.9 million forecasted observations, TS-Arena is planned to be expand to additional domains, multi-variate forecasting, and probabilistic evaluation, while obtaining clear protocols for preventing information leakage and inflated performance measures. In addition, we release a quarterly updated archive on HuggingFace as a offline mirror of the live leaderboard, enabling fast independent evaluation of new models.

\bibliographystyle{ACM-Reference-Format}
\bibliography{sample-base}

\appendix
\section{Detailed Experimental Setup}
The TS-Arena evaluation infrastructure is implemented as a three-tier microservices architecture orchestrated via Docker Compose. This design ensures modularity and reproducibility by isolating distinct functional layers: the Challenge Upload Service, the Master Controller API, and the Model Service Layer. All components communicate over an isolated internal Docker bridge network, with the Master Controller acting as the sole gateway. Users register on the platform to obtain an API upload key and user ID, stored locally in an environment file. All external communication with the TS-Arena platform is authenticated via the API-key, while internal routing relies on a strict separation of concerns between the orchestration and forecasting layers.

The core orchestration logic resides in the Master Controller, which exposes a standard /predict endpoint via FastAPI. To manage the high resource demands of TSFMs, the system employs an on-demand Worker pattern. Rather than keeping all model containers active, the Master Controller dynamically manages container lifecycles. It interacts with the Docker daemon through a restricted Docker Socket Proxy.
When a prediction request arrives, the Worker class identifies the target container by its registered name. It starts the container and enters a polling loop, sending repeated requests to the container’s /health endpoint. A default timeout of 300 seconds accommodates the possible initialization latency of TSFMs. Once the model returns a 200 OK status, the payload is transmitted. Immediately following the inference response, the container is stopped to free GPU resources. This distinct lifecycle management allows the evaluation of multiple GPU-heavy models on hardware with limited GPU memory.

To support a heterogeneous mix of forecasting approaches, every model service adheres to a unified directory structure and interface contract. Each service consists of a Dockerfile, a requirements file for dependency management, and an application directory separating HTTP handling (main.py) from prediction logic (model.py). The main.py module implements a FastAPI application with two mandatory endpoints: a GET /health probe and a POST /predict interface. The prediction endpoint accepts historical time series (in single or batch format), a forecast horizon, and a frequency string.
This abstraction layer normalizes the varying input requirements of the underlying models. For instance, models like Chronos-2 and Moirai require data conversion into Pandas DataFrames, whereas deep learning baselines may require normalized Float tensors. The service layer also handles timestamp reconstruction. Since models output raw numerical values, the service calculates future timestamps based on the input frequency (e.g., 15min, D, M). Furthermore, the services standardize probabilistic output. Point forecasts are derived from the median (0.5 quantile), while probabilistic values are returned as a dictionary of quantiles (e.g., 0.1, 0.2, ..., 0.8, 0.9) if the model natively allows for that.

The system supports two distinct categories of forecasting models, managed via a JSON configuration file that maps local Docker container names to their platform-registered identities:
\begin{itemize}
    \item Pretrained TSFMs: Services such as Chronos, TimesFM, Moirai, VisionTS, and TinyTimeMixer leverage pretrained weights. These weights are loaded at startup, either from the HuggingFace Hub or local caches. To optimize bandwidth and startup time, a shared Docker volume persists weights across container restarts. The infrastructure supports multiple loading options, including the transformers library, model-specific pip packages (e.g., chronos-forecasting), and manual checkpoint downloads.
    \item Statistical Baselines: Pure Python implementations such as Seasonal Naive and Simple Moving Average are deployed in lightweight containers without GPU requirements.
\end{itemize}

The Challenge Upload Service is responsible for the evaluation loop. It executes a continuous polling cycle, fetching available challenge rounds from the API. Upon detecting a new round, it parses the ISO 8601 duration strings (e.g., PT1H) to determine step frequency and horizon length. The service retrieves the historical context and resolves which local containers correspond to the active models registered in the user’s account. It then dispatches prediction requests to the Master Controller. To ensure robust operation, the service maintains a participation log CSV, recording the timestamp, round ID, challenge name, container name, API model name, success/failure status, and error details. Docker Compose profiles allow users to selectively build and deploy specific model families, while overlay files manage hardware-specific configurations, such as NVIDIA GPU reservations or platform emulation for Apple Silicon.

\section{Selection of Time Series}
\label{sec:ts_selection}

\begin{figure}[H]
    \centering
    \includegraphics[width=1\linewidth]{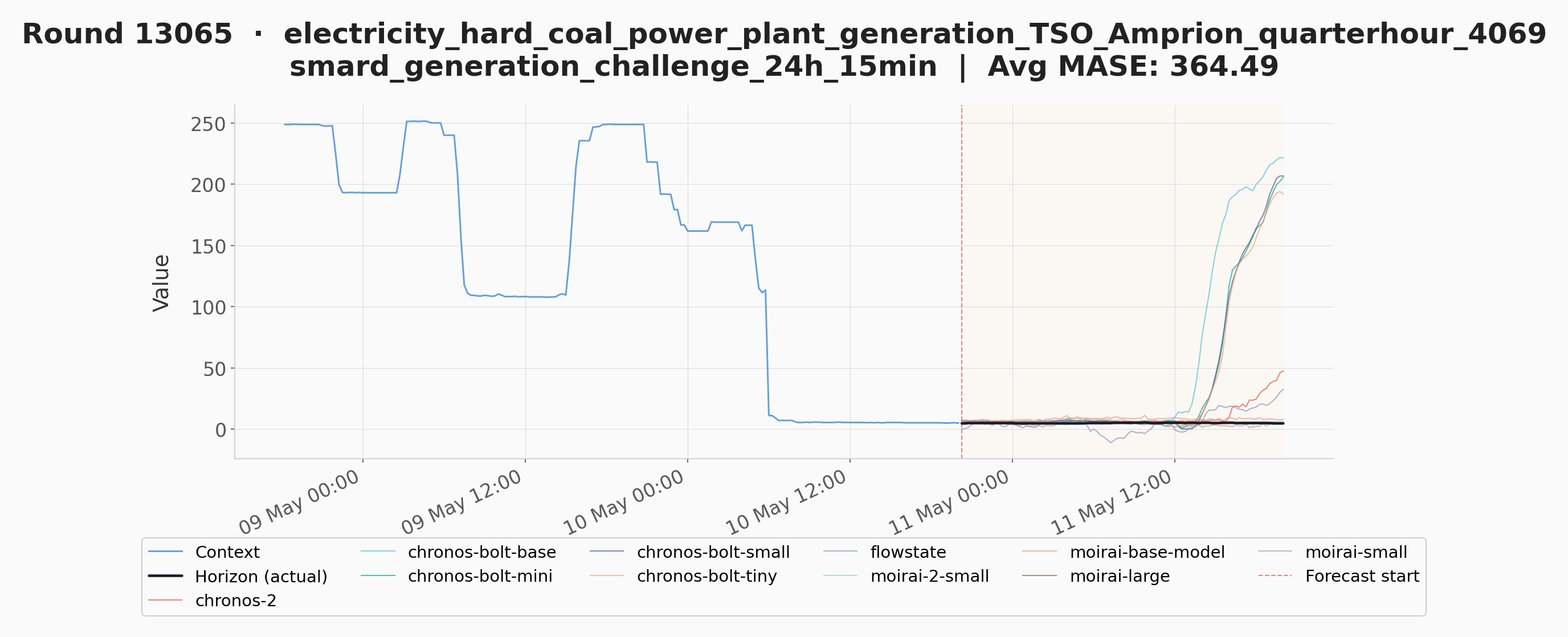}
    \caption{Example of a hard coal challenge with high average MASE}
    \label{fig:hard_coal_single}
\end{figure}

\begin{figure}[H]
    \centering
    \includegraphics[width=1\linewidth]{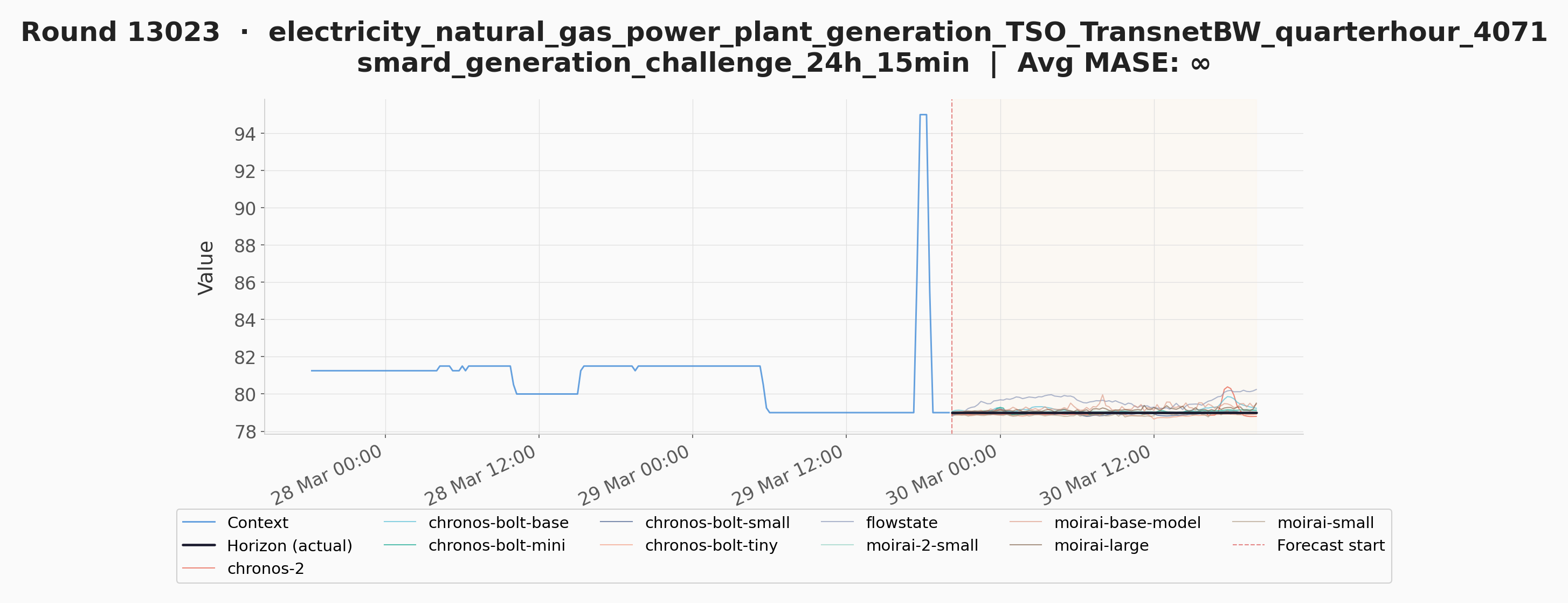}
    \caption{Example of a natural gas challenge with high average MASE}
    \label{fig:nat_gas_single}
\end{figure}

\begin{figure}[H]
    \centering
    \includegraphics[width=1\linewidth]{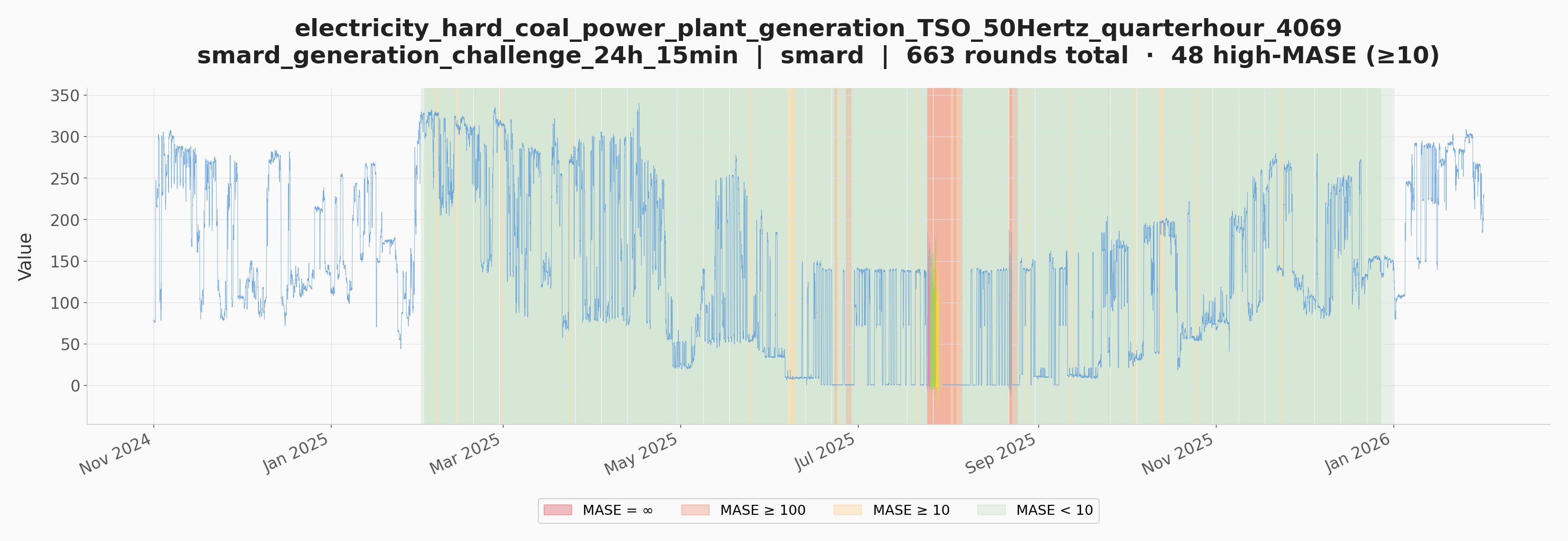}
    \caption{Example of a complete hard coal time series with many high average MASE rounds}
    \label{fig:hard_coal_comb}
\end{figure}

\begin{figure}[H]
    \centering
    \includegraphics[width=1\linewidth]{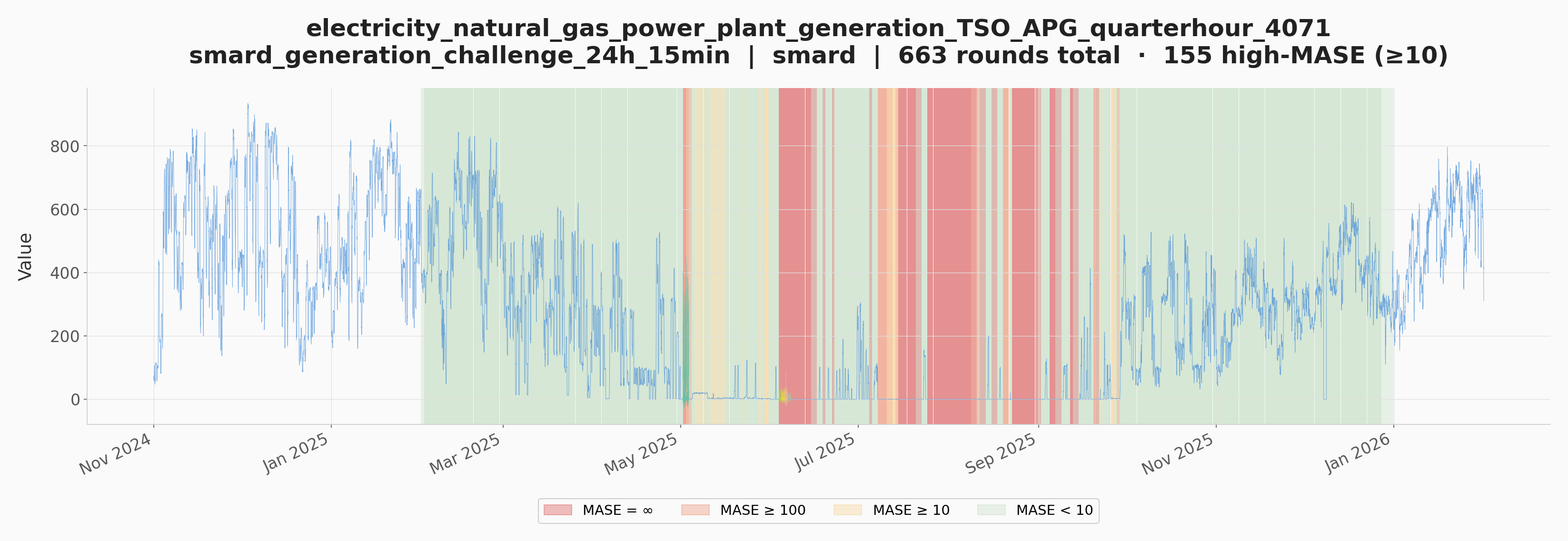}
    \caption{Example of a complete natural gas time series with many high average MASE rounds}
    \label{fig:nat_gas_comb}
\end{figure}

As can be seen in \autoref{fig:hard_coal_single} and \autoref{fig:nat_gas_single}, both hard coal and natural gas exhibit patterns that are constant for a large amount of time, making it hard for univariate forecasting models to achieve accurate predictions. This results in a very high average MASE across the models. \autoref{fig:hard_coal_comb} and \autoref{fig:nat_gas_comb} further illustrate that, showing a seemingly high weather dependency. In the warmer months, the demand is mostly covered by the renewable energy sources, leading to hard coal and natural gas hardly being used. This is also the time when challenges would result in very inaccurate predictions and high MASE values.


\begin{table*}[b!]
\centering
\small
\begin{tabular}{llcccccc}
\toprule
& & \multicolumn{2}{c}{\textbf{price - 15min - 24h}} & \multicolumn{2}{c}{\textbf{generation - 15min - 24h}} & \multicolumn{2}{c}{\textbf{consumption - 15min - 24h}} \\
\cmidrule(lr){3-4} \cmidrule(lr){5-6} \cmidrule(lr){7-8}
Model & Rounds & ELO$^{(\text{CI})}$ & MASE$_{(\pm\text{std})}$ & ELO$^{(\text{CI})}$ & MASE$_{(\pm\text{std})}$ & ELO$^{(\text{CI})}$ & MASE$_{(\pm\text{std})}$ \\
\midrule
tirex & 650 & $\mathbf{1283^{+75}_{-78}}$ & $\mathbf{0.706}_{{\pm 1.00}}$ & $\mathbf{1354^{+74}_{-70}}$ & $\mathbf{0.852}_{{\pm 1.55}}$ & $1358^{+66}_{-56}$ & $0.272_{{\pm 0.20}}$ \\
chronos-bolt-base & 1095 & $1245^{+83}_{-91}$ & $0.747_{{\pm 1.50}}$ & $1202^{+83}_{-80}$ & $0.969_{{\pm 2.10}}$ & $\mathbf{1499^{+64}_{-66}}$ & $0.259_{{\pm 0.21}}$ \\
chronos-2 & 228 & $1244^{+73}_{-83}$ & $0.768_{{\pm 0.86}}$ & $1267^{+83}_{-86}$ & $1.020_{{\pm 2.59}}$ & $1423^{+66}_{-67}$ & $0.247_{{\pm 0.19}}$ \\
flowstate & 441 & $1232^{+99}_{-98}$ & $0.765_{{\pm 1.24}}$ & $1226^{+98}_{-101}$ & $0.956_{{\pm 2.43}}$ & $1453^{+70}_{-70}$ & $\mathbf{0.229}_{{\pm 0.19}}$ \\
timesfm-2.5-200m & 323 & $1225^{+81}_{-95}$ & $0.843_{{\pm 1.92}}$ & $1302^{+61}_{-82}$ & $1.005_{{\pm 1.99}}$ & $1444^{+56}_{-68}$ & $0.233_{{\pm 0.19}}$ \\
sundial-128m & 999 & $1215^{+81}_{-83}$ & $0.756_{{\pm 1.38}}$ & $1166^{+68}_{-74}$ & $0.961_{{\pm 1.90}}$ & $1349^{+65}_{-57}$ & $0.294_{{\pm 0.23}}$ \\
timesfm-2.0-500m & 1094 & $1201^{+75}_{-89}$ & $0.768_{{\pm 1.34}}$ & $1174^{+67}_{-82}$ & $0.986_{{\pm 1.82}}$ & $1325^{+75}_{-68}$ & $0.302_{{\pm 0.24}}$ \\
chronos-bolt-small & 1095 & $1184^{+91}_{-97}$ & $0.778_{{\pm 1.52}}$ & $1210^{+78}_{-82}$ & $0.958_{{\pm 2.09}}$ & $1438^{+70}_{-75}$ & $0.269_{{\pm 0.22}}$ \\
moirai-2-small & 150 & $1164^{+78}_{-85}$ & $0.846_{{\pm 1.05}}$ & $1311^{+70}_{-84}$ & $1.033_{{\pm 3.41}}$ & $1362^{+59}_{-63}$ & $0.252_{{\pm 0.20}}$ \\
chronos-bolt-mini & 1095 & $1162^{+91}_{-93}$ & $0.786_{{\pm 1.33}}$ & $1185^{+79}_{-87}$ & $0.961_{{\pm 1.99}}$ & $1426^{+63}_{-68}$ & $0.268_{{\pm 0.21}}$ \\
tinytimemixer-r2-1024-96 & 999 & $1156^{+72}_{-72}$ & $0.776_{{\pm 1.47}}$ & $1035^{+87}_{-77}$ & $1.016_{{\pm 1.91}}$ & $1124^{+68}_{-70}$ & $0.403_{{\pm 0.27}}$ \\
chronos-bolt-tiny & 1095 & $1133^{+99}_{-95}$ & $0.807_{{\pm 1.31}}$ & $1182^{+82}_{-80}$ & $0.964_{{\pm 1.94}}$ & $1389^{+77}_{-78}$ & $0.274_{{\pm 0.22}}$ \\
time-moe-200m & 1094 & $1129^{+81}_{-81}$ & $0.832_{{\pm 2.03}}$ & $993^{+86}_{-74}$ & $1.139_{{\pm 2.26}}$ & $1099^{+73}_{-62}$ & $0.375_{{\pm 0.24}}$ \\
time-moe-50m & 1095 & $1113^{+83}_{-94}$ & $0.847_{{\pm 2.09}}$ & $970^{+85}_{-73}$ & $1.155_{{\pm 2.20}}$ & $1045^{+70}_{-73}$ & $0.394_{{\pm 0.25}}$ \\
tinytimemixer-r1-1024-96 & 1094 & $1032^{+86}_{-70}$ & $0.845_{{\pm 1.29}}$ & $1069^{+84}_{-89}$ & $0.975_{{\pm 1.96}}$ & $985^{+58}_{-57}$ & $0.443_{{\pm 0.24}}$ \\
seasonal-naive & 1095 & $996^{+121}_{-126}$ & $0.930_{{\pm 1.67}}$ & $896^{+106}_{-102}$ & $1.290_{{\pm 2.30}}$ & $928^{+121}_{-121}$ & $0.643_{{\pm 0.49}}$ \\
moirai-base-model & 1095 & $975^{+72}_{-70}$ & $0.852_{{\pm 0.85}}$ & $1218^{+95}_{-94}$ & $0.877_{{\pm 0.68}}$ & $975^{+69}_{-63}$ & $0.489_{{\pm 0.32}}$ \\
moirai-large & 1095 & $966^{+90}_{-91}$ & $0.866_{{\pm 0.87}}$ & $1211^{+92}_{-92}$ & $0.910_{{\pm 1.26}}$ & $835^{+57}_{-51}$ & $0.563_{{\pm 0.34}}$ \\
moirai-small & 1095 & $945^{+90}_{-84}$ & $0.870_{{\pm 0.87}}$ & $1154^{+96}_{-99}$ & $0.930_{{\pm 0.90}}$ & $873^{+67}_{-51}$ & $0.535_{{\pm 0.30}}$ \\
seasonal-average & 1095 & $914^{+102}_{-103}$ & $1.061_{{\pm 2.59}}$ & $747^{+99}_{-79}$ & $1.540_{{\pm 3.40}}$ & $665^{+121}_{-104}$ & $0.858_{{\pm 0.58}}$ \\
visiontspp-large & 443 & $856^{+100}_{-76}$ & $1.120_{{\pm 3.36}}$ & $888^{+112}_{-87}$ & $1.464_{{\pm 2.80}}$ & $617^{+36}_{-36}$ & $0.701_{{\pm 0.35}}$ \\
visiontspp-base & 443 & $804^{+103}_{-78}$ & $1.142_{{\pm 3.26}}$ & $1008^{+96}_{-83}$ & $1.200_{{\pm 1.94}}$ & $675^{+45}_{-35}$ & $0.636_{{\pm 0.33}}$ \\
simple-moving-average & 1095 & $660^{+87}_{-70}$ & $1.306_{{\pm 2.77}}$ & $474^{+66}_{-53}$ & $2.215_{{\pm 4.70}}$ & $414^{+26}_{-26}$ & $1.089_{{\pm 0.40}}$ \\
\bottomrule
\end{tabular}
\caption{ELO Leaderboard: SMARD (15min/24h)}
\label{tab:elo_smard_15min}
\end{table*}

\begin{table*}[b!]
\centering
\small
\begin{tabular}{llcccccc}
\toprule
& & \multicolumn{2}{c}{\textbf{price - 1h - 72h}} & \multicolumn{2}{c}{\textbf{generation - 1h - 72h}} & \multicolumn{2}{c}{\textbf{consumption - 1h - 72h}} \\
\cmidrule(lr){3-4} \cmidrule(lr){5-6} \cmidrule(lr){7-8}
Model & Rounds & ELO$^{(\text{CI})}$ & MASE$_{(\pm\text{std})}$ & ELO$^{(\text{CI})}$ & MASE$_{(\pm\text{std})}$ & ELO$^{(\text{CI})}$ & MASE$_{(\pm\text{std})}$ \\
\midrule
tirex & 651 & $\mathbf{1340^{+80}_{-89}}$ & $\mathbf{0.646}_{{\pm 0.32}}$ & $1331^{+75}_{-91}$ & $\mathbf{0.756}_{{\pm 0.59}}$ & $1260^{+84}_{-86}$ & $0.267_{{\pm 0.19}}$ \\
chronos-2 & 228 & $1328^{+78}_{-89}$ & $0.720_{{\pm 0.39}}$ & $\mathbf{1362^{+69}_{-66}}$ & $0.851_{{\pm 0.65}}$ & $\mathbf{1515^{+53}_{-55}}$ & $\mathbf{0.224}_{{\pm 0.17}}$ \\
timesfm-2.0-500m & 1095 & $1271^{+68}_{-73}$ & $0.671_{{\pm 0.34}}$ & $1238^{+75}_{-77}$ & $0.775_{{\pm 0.64}}$ & $1382^{+73}_{-87}$ & $0.248_{{\pm 0.18}}$ \\
moirai-2-small & 150 & $1241^{+61}_{-66}$ & $0.752_{{\pm 0.42}}$ & $1224^{+65}_{-66}$ & $0.881_{{\pm 0.59}}$ & $1434^{+54}_{-60}$ & $0.248_{{\pm 0.20}}$ \\
chronos-bolt-base & 1095 & $1238^{+87}_{-95}$ & $0.688_{{\pm 0.37}}$ & $1083^{+100}_{-93}$ & $0.887_{{\pm 1.10}}$ & $1421^{+61}_{-76}$ & $0.243_{{\pm 0.17}}$ \\
flowstate & 441 & $1228^{+97}_{-115}$ & $0.720_{{\pm 0.38}}$ & $1313^{+75}_{-87}$ & $0.814_{{\pm 0.68}}$ & $1300^{+72}_{-80}$ & $0.239_{{\pm 0.17}}$ \\
time-moe-50m & 1095 & $1215^{+81}_{-88}$ & $0.685_{{\pm 0.39}}$ & $1049^{+102}_{-87}$ & $0.847_{{\pm 0.87}}$ & $1119^{+81}_{-84}$ & $0.303_{{\pm 0.19}}$ \\
sundial-128m & 999 & $1194^{+78}_{-78}$ & $0.683_{{\pm 0.39}}$ & $1216^{+73}_{-78}$ & $0.767_{{\pm 0.69}}$ & $997^{+90}_{-76}$ & $0.342_{{\pm 0.22}}$ \\
chronos-bolt-small & 1095 & $1192^{+97}_{-86}$ & $0.709_{{\pm 0.45}}$ & $1150^{+72}_{-85}$ & $0.843_{{\pm 0.97}}$ & $1401^{+62}_{-62}$ & $0.243_{{\pm 0.16}}$ \\
chronos-bolt-mini & 1095 & $1192^{+84}_{-92}$ & $0.702_{{\pm 0.41}}$ & $1179^{+79}_{-84}$ & $0.832_{{\pm 0.95}}$ & $1429^{+69}_{-72}$ & $0.238_{{\pm 0.16}}$ \\
chronos-bolt-tiny & 1095 & $1182^{+69}_{-84}$ & $0.700_{{\pm 0.36}}$ & $1210^{+77}_{-81}$ & $0.814_{{\pm 0.84}}$ & $1368^{+75}_{-79}$ & $0.244_{{\pm 0.16}}$ \\
moirai-large & 1095 & $1171^{+79}_{-75}$ & $0.699_{{\pm 0.35}}$ & $1242^{+74}_{-87}$ & $0.771_{{\pm 0.67}}$ & $1245^{+66}_{-75}$ & $0.282_{{\pm 0.20}}$ \\
time-moe-200m & 1095 & $1148^{+81}_{-78}$ & $0.707_{{\pm 0.40}}$ & $1064^{+83}_{-77}$ & $0.846_{{\pm 0.87}}$ & $1200^{+78}_{-77}$ & $0.285_{{\pm 0.18}}$ \\
tinytimemixer-r2-1024-96 & 992 & $1113^{+84}_{-73}$ & $0.716_{{\pm 0.40}}$ & $1023^{+94}_{-75}$ & $0.841_{{\pm 0.86}}$ & $1012^{+72}_{-61}$ & $0.316_{{\pm 0.16}}$ \\
timesfm-2.5-200m & 324 & $1111^{+70}_{-74}$ & $0.786_{{\pm 0.41}}$ & $1297^{+59}_{-67}$ & $0.912_{{\pm 0.86}}$ & $1237^{+79}_{-78}$ & $0.257_{{\pm 0.19}}$ \\
moirai-base-model & 1095 & $1104^{+68}_{-66}$ & $0.720_{{\pm 0.35}}$ & $1264^{+70}_{-84}$ & $0.764_{{\pm 0.66}}$ & $1233^{+73}_{-77}$ & $0.280_{{\pm 0.19}}$ \\
tinytimemixer-r1-1024-96 & 1095 & $1061^{+78}_{-78}$ & $0.745_{{\pm 0.44}}$ & $1056^{+87}_{-74}$ & $0.837_{{\pm 0.83}}$ & $1070^{+83}_{-73}$ & $0.316_{{\pm 0.18}}$ \\
moirai-small & 1095 & $960^{+77}_{-70}$ & $0.784_{{\pm 0.34}}$ & $1179^{+91}_{-105}$ & $0.777_{{\pm 0.54}}$ & $1004^{+68}_{-59}$ & $0.332_{{\pm 0.21}}$ \\
seasonal-naive & 1095 & $878^{+134}_{-105}$ & $0.920_{{\pm 0.56}}$ & $973^{+122}_{-115}$ & $0.925_{{\pm 0.80}}$ & $683^{+131}_{-108}$ & $0.669_{{\pm 0.35}}$ \\
seasonal-average & 1095 & $876^{+104}_{-93}$ & $0.922_{{\pm 0.64}}$ & $942^{+103}_{-102}$ & $0.952_{{\pm 1.14}}$ & $629^{+84}_{-71}$ & $0.703_{{\pm 0.31}}$ \\
visiontspp-large & 444 & $728^{+82}_{-61}$ & $1.031_{{\pm 0.38}}$ & $729^{+80}_{-70}$ & $1.285_{{\pm 1.25}}$ & $528^{+38}_{-39}$ & $0.834_{{\pm 0.28}}$ \\
simple-moving-average & 1095 & $668^{+72}_{-59}$ & $1.086_{{\pm 0.52}}$ & $466^{+50}_{-36}$ & $1.365_{{\pm 1.38}}$ & $463^{+39}_{-34}$ & $0.863_{{\pm 0.31}}$ \\
visiontspp-base & 444 & $651^{+86}_{-67}$ & $1.089_{{\pm 0.43}}$ & $806^{+87}_{-74}$ & $1.225_{{\pm 1.11}}$ & $551^{+30}_{-33}$ & $0.818_{{\pm 0.26}}$ \\
\bottomrule
\end{tabular}
\caption{ELO Leaderboard: SMARD (1h/72h)}
\label{tab:elo_smard_1h}
\end{table*}

\begin{table*}[h!]
\centering
\small
\begin{tabular}{llcccccc}
\toprule
& & \multicolumn{2}{c}{\textbf{generation - 15min - 24h}} & \multicolumn{2}{c}{\textbf{price - 15min - 24h}} & \multicolumn{2}{c}{\textbf{consumption - 15min - 24h}} \\
\cmidrule(lr){3-4} \cmidrule(lr){5-6} \cmidrule(lr){7-8}
Model & Rounds & ELO$^{(\text{CI})}$ & MASE$_{(\pm\text{std})}$ & ELO$^{(\text{CI})}$ & MASE$_{(\pm\text{std})}$ & ELO$^{(\text{CI})}$ & MASE$_{(\pm\text{std})}$ \\
\midrule
chronos-2 & 280 & $\mathbf{1424^{+70}_{-70}}$ & $0.635_{{\pm 0.52}}$ & $1161^{+76}_{-77}$ & $0.890_{{\pm 0.35}}$ & $1296^{+69}_{-81}$ & $0.401_{{\pm 0.28}}$ \\
timesfm-2.5-200m & 375 & $1419^{+69}_{-76}$ & $0.642_{{\pm 0.55}}$ & $1174^{+74}_{-83}$ & $0.850_{{\pm 0.36}}$ & $1285^{+72}_{-71}$ & $0.407_{{\pm 0.26}}$ \\
tirex & 702 & $1408^{+61}_{-66}$ & $\mathbf{0.618}_{{\pm 0.75}}$ & $1173^{+77}_{-89}$ & $0.849_{{\pm 0.35}}$ & $1265^{+63}_{-67}$ & $0.461_{{\pm 0.34}}$ \\
moirai-2-small & 202 & $1390^{+57}_{-62}$ & $0.653_{{\pm 0.45}}$ & $1042^{+86}_{-87}$ & $1.075_{{\pm 0.56}}$ & $1306^{+66}_{-78}$ & $\mathbf{0.389}_{{\pm 0.25}}$ \\
chronos-bolt-small & 1085 & $1296^{+82}_{-84}$ & $0.653_{{\pm 0.73}}$ & $1177^{+89}_{-82}$ & $0.882_{{\pm 0.49}}$ & $1275^{+75}_{-86}$ & $0.494_{{\pm 0.38}}$ \\
chronos-bolt-base & 1085 & $1290^{+70}_{-97}$ & $0.645_{{\pm 0.74}}$ & $\mathbf{1194^{+78}_{-79}}$ & $0.861_{{\pm 0.42}}$ & $\mathbf{1328^{+73}_{-75}}$ & $0.482_{{\pm 0.37}}$ \\
chronos-bolt-tiny & 1085 & $1270^{+87}_{-86}$ & $0.644_{{\pm 0.62}}$ & $1176^{+83}_{-89}$ & $0.871_{{\pm 0.45}}$ & $1216^{+100}_{-97}$ & $0.507_{{\pm 0.39}}$ \\
chronos-bolt-mini & 1085 & $1268^{+73}_{-82}$ & $0.644_{{\pm 0.65}}$ & $1179^{+76}_{-90}$ & $0.882_{{\pm 0.49}}$ & $1250^{+87}_{-86}$ & $0.498_{{\pm 0.39}}$ \\
moirai-base-model & 1085 & $1232^{+82}_{-80}$ & $0.684_{{\pm 0.49}}$ & $1020^{+89}_{-88}$ & $0.939_{{\pm 0.34}}$ & $946^{+84}_{-76}$ & $0.634_{{\pm 0.30}}$ \\
flowstate & 493 & $1219^{+91}_{-89}$ & $0.636_{{\pm 0.68}}$ & $1148^{+107}_{-106}$ & $\mathbf{0.839}_{{\pm 0.34}}$ & $1318^{+86}_{-93}$ & $0.406_{{\pm 0.30}}$ \\
sundial-128m & 1050 & $1180^{+75}_{-78}$ & $0.676_{{\pm 0.81}}$ & $1158^{+69}_{-73}$ & $0.908_{{\pm 0.52}}$ & $1263^{+81}_{-74}$ & $0.507_{{\pm 0.38}}$ \\
timesfm-2.0-500m & 1084 & $1166^{+77}_{-73}$ & $0.719_{{\pm 0.82}}$ & $1125^{+80}_{-87}$ & $0.908_{{\pm 0.50}}$ & $1200^{+70}_{-77}$ & $0.534_{{\pm 0.44}}$ \\
moirai-large & 1085 & $1131^{+84}_{-96}$ & $0.745_{{\pm 0.69}}$ & $999^{+84}_{-86}$ & $0.960_{{\pm 0.39}}$ & $901^{+86}_{-81}$ & $0.657_{{\pm 0.33}}$ \\
moirai-small & 1085 & $1123^{+82}_{-90}$ & $0.748_{{\pm 0.63}}$ & $1005^{+92}_{-88}$ & $0.941_{{\pm 0.33}}$ & $838^{+94}_{-94}$ & $0.702_{{\pm 0.28}}$ \\
tinytimemixer-r1-1024-96 & 1084 & $1041^{+89}_{-81}$ & $0.679_{{\pm 0.62}}$ & $1054^{+87}_{-98}$ & $1.035_{{\pm 0.78}}$ & $1152^{+89}_{-80}$ & $0.535_{{\pm 0.31}}$ \\
tinytimemixer-r2-1024-96 & 1084 & $983^{+88}_{-84}$ & $0.696_{{\pm 0.78}}$ & $1061^{+99}_{-96}$ & $1.026_{{\pm 0.75}}$ & $1233^{+79}_{-74}$ & $0.518_{{\pm 0.36}}$ \\
visiontspp-base & 495 & $965^{+70}_{-65}$ & $0.817_{{\pm 0.81}}$ & $910^{+95}_{-89}$ & $1.083_{{\pm 0.88}}$ & $850^{+66}_{-63}$ & $0.618_{{\pm 0.34}}$ \\
seasonal-naive & 1093 & $941^{+91}_{-95}$ & $0.880_{{\pm 1.17}}$ & $853^{+114}_{-100}$ & $1.203_{{\pm 0.97}}$ & $1029^{+120}_{-105}$ & $0.606_{{\pm 0.46}}$ \\
time-moe-50m & 1084 & $896^{+75}_{-57}$ & $0.806_{{\pm 1.15}}$ & $1071^{+84}_{-89}$ & $0.998_{{\pm 0.65}}$ & $1039^{+80}_{-59}$ & $0.584_{{\pm 0.38}}$ \\
time-moe-200m & 1084 & $888^{+83}_{-66}$ & $0.794_{{\pm 1.15}}$ & $1058^{+90}_{-87}$ & $1.002_{{\pm 0.64}}$ & $910^{+85}_{-72}$ & $0.654_{{\pm 0.41}}$ \\
visiontspp-large & 495 & $841^{+78}_{-61}$ & $0.943_{{\pm 1.16}}$ & $873^{+98}_{-81}$ & $1.125_{{\pm 0.97}}$ & $808^{+69}_{-69}$ & $0.653_{{\pm 0.36}}$ \\
seasonal-average & 1091 & $724^{+86}_{-78}$ & $1.032_{{\pm 1.60}}$ & $872^{+123}_{-94}$ & $1.381_{{\pm 2.07}}$ & $978^{+105}_{-113}$ & $0.661_{{\pm 0.51}}$ \\
simple-moving-average & 1084 & $381^{+27}_{-30}$ & $1.453_{{\pm 2.25}}$ & $827^{+105}_{-103}$ & $1.470_{{\pm 1.47}}$ & $550^{+72}_{-52}$ & $0.910_{{\pm 0.41}}$ \\
\bottomrule
\end{tabular}
\caption{ELO Leaderboard: gridstatus (15min/24h)}
\label{tab:elo_gridstatus_15min}
\end{table*}

\begin{table*}[h!]
\centering
\small
\begin{tabular}{llcccccc}
\toprule
& & \multicolumn{2}{c}{\textbf{generation - 1h - 72h}} & \multicolumn{2}{c}{\textbf{price - 1h - 72h}} & \multicolumn{2}{c}{\textbf{consumption - 1h - 72h}} \\
\cmidrule(lr){3-4} \cmidrule(lr){5-6} \cmidrule(lr){7-8}
Model & Rounds & ELO$^{(\text{CI})}$ & MASE$_{(\pm\text{std})}$ & ELO$^{(\text{CI})}$ & MASE$_{(\pm\text{std})}$ & ELO$^{(\text{CI})}$ & MASE$_{(\pm\text{std})}$ \\
\midrule
tirex & 736 & $\mathbf{1473^{+63}_{-69}}$ & $\mathbf{0.636}_{{\pm 0.61}}$ & $1137^{+88}_{-79}$ & $\mathbf{0.825}_{{\pm 0.31}}$ & $1248^{+74}_{-75}$ & $0.579_{{\pm 0.33}}$ \\
timesfm-2.5-200m & 409 & $1413^{+76}_{-77}$ & $0.691_{{\pm 0.73}}$ & $1116^{+79}_{-93}$ & $0.872_{{\pm 0.36}}$ & $1192^{+87}_{-99}$ & $0.565_{{\pm 0.31}}$ \\
moirai-2-small & 235 & $1397^{+56}_{-61}$ & $0.700_{{\pm 0.91}}$ & $1107^{+83}_{-87}$ & $0.959_{{\pm 0.35}}$ & $\mathbf{1269^{+82}_{-85}}$ & $0.537_{{\pm 0.28}}$ \\
chronos-2 & 313 & $1395^{+71}_{-91}$ & $0.699_{{\pm 0.84}}$ & $\mathbf{1148^{+86}_{-103}}$ & $0.886_{{\pm 0.33}}$ & $1266^{+75}_{-79}$ & $\mathbf{0.534}_{{\pm 0.30}}$ \\
moirai-base-model & 1087 & $1306^{+81}_{-94}$ & $0.676_{{\pm 0.62}}$ & $1082^{+86}_{-86}$ & $0.880_{{\pm 0.35}}$ & $1130^{+74}_{-74}$ & $0.640_{{\pm 0.33}}$ \\
moirai-large & 1087 & $1269^{+77}_{-81}$ & $0.688_{{\pm 0.65}}$ & $1085^{+80}_{-81}$ & $0.869_{{\pm 0.33}}$ & $1142^{+83}_{-82}$ & $0.636_{{\pm 0.32}}$ \\
chronos-bolt-small & 1087 & $1256^{+72}_{-82}$ & $0.658_{{\pm 0.66}}$ & $1118^{+80}_{-95}$ & $0.873_{{\pm 0.39}}$ & $1235^{+84}_{-79}$ & $0.601_{{\pm 0.33}}$ \\
chronos-bolt-base & 1087 & $1232^{+81}_{-82}$ & $0.669_{{\pm 0.73}}$ & $1146^{+89}_{-94}$ & $0.853_{{\pm 0.40}}$ & $1247^{+83}_{-92}$ & $0.598_{{\pm 0.34}}$ \\
chronos-bolt-tiny & 1087 & $1232^{+73}_{-69}$ & $0.648_{{\pm 0.65}}$ & $1123^{+85}_{-96}$ & $0.851_{{\pm 0.35}}$ & $1232^{+77}_{-82}$ & $0.603_{{\pm 0.32}}$ \\
chronos-bolt-mini & 1087 & $1216^{+70}_{-77}$ & $0.652_{{\pm 0.65}}$ & $1126^{+88}_{-96}$ & $0.861_{{\pm 0.37}}$ & $1247^{+79}_{-67}$ & $0.598_{{\pm 0.33}}$ \\
timesfm-2.0-500m & 1087 & $1192^{+104}_{-111}$ & $0.688_{{\pm 0.68}}$ & $1108^{+84}_{-89}$ & $0.859_{{\pm 0.40}}$ & $1189^{+87}_{-97}$ & $0.630_{{\pm 0.44}}$ \\
moirai-small & 1087 & $1184^{+80}_{-99}$ & $0.710_{{\pm 0.59}}$ & $1032^{+90}_{-94}$ & $0.887_{{\pm 0.31}}$ & $1022^{+79}_{-67}$ & $0.675_{{\pm 0.33}}$ \\
flowstate & 526 & $1121^{+85}_{-86}$ & $0.665_{{\pm 0.72}}$ & $1143^{+92}_{-103}$ & $0.854_{{\pm 0.33}}$ & $1267^{+103}_{-105}$ & $0.566_{{\pm 0.34}}$ \\
sundial-128m & 1084 & $1121^{+73}_{-66}$ & $0.688_{{\pm 0.72}}$ & $1116^{+81}_{-90}$ & $0.895_{{\pm 0.44}}$ & $1143^{+83}_{-79}$ & $0.629_{{\pm 0.35}}$ \\
seasonal-naive & 1094 & $1113^{+99}_{-111}$ & $0.779_{{\pm 0.78}}$ & $848^{+121}_{-105}$ & $1.095_{{\pm 0.69}}$ & $942^{+123}_{-101}$ & $0.745_{{\pm 0.39}}$ \\
seasonal-average & 1092 & $1011^{+102}_{-104}$ & $0.813_{{\pm 0.92}}$ & $906^{+110}_{-99}$ & $1.101_{{\pm 0.91}}$ & $979^{+105}_{-110}$ & $0.740_{{\pm 0.41}}$ \\
time-moe-50m & 1087 & $882^{+68}_{-64}$ & $0.743_{{\pm 0.90}}$ & $1096^{+89}_{-100}$ & $0.931_{{\pm 0.49}}$ & $1184^{+91}_{-94}$ & $0.625_{{\pm 0.36}}$ \\
tinytimemixer-r1-1024-96 & 1087 & $876^{+85}_{-74}$ & $0.691_{{\pm 0.65}}$ & $1061^{+88}_{-91}$ & $0.955_{{\pm 0.55}}$ & $1142^{+93}_{-86}$ & $0.632_{{\pm 0.32}}$ \\
visiontspp-base & 529 & $872^{+65}_{-56}$ & $0.981_{{\pm 0.75}}$ & $887^{+101}_{-84}$ & $1.067_{{\pm 0.55}}$ & $641^{+63}_{-48}$ & $1.005_{{\pm 0.35}}$ \\
time-moe-200m & 1087 & $851^{+72}_{-59}$ & $0.744_{{\pm 0.88}}$ & $1076^{+109}_{-93}$ & $0.935_{{\pm 0.51}}$ & $1179^{+85}_{-91}$ & $0.625_{{\pm 0.36}}$ \\
tinytimemixer-r2-1024-96 & 1087 & $817^{+88}_{-69}$ & $0.705_{{\pm 0.71}}$ & $1071^{+91}_{-97}$ & $0.966_{{\pm 0.56}}$ & $1128^{+83}_{-83}$ & $0.634_{{\pm 0.31}}$ \\
visiontspp-large & 529 & $771^{+71}_{-77}$ & $1.071_{{\pm 0.98}}$ & $879^{+119}_{-92}$ & $1.076_{{\pm 0.58}}$ & $629^{+65}_{-47}$ & $1.025_{{\pm 0.36}}$ \\
simple-moving-average & 1087 & $360^{+34}_{-38}$ & $1.343_{{\pm 1.79}}$ & $816^{+128}_{-94}$ & $1.381_{{\pm 1.12}}$ & $555^{+71}_{-53}$ & $1.120_{{\pm 0.49}}$ \\
\bottomrule
\end{tabular}
\caption{ELO Leaderboard: gridstatus (1h/72h)}
\label{tab:elo_gridstatus_1h}
\end{table*}

\begin{table*}[ht]
\centering
\small
\begin{tabular}{llcccc}
\toprule
& & \multicolumn{2}{c}{\textbf{challenge - 15min - 24h}} & \multicolumn{2}{c}{\textbf{challenge - 1h - 72h}} \\
\cmidrule(lr){3-4} \cmidrule(lr){5-6}
Model & Rounds & ELO$^{(\text{CI})}$ & MASE$_{(\pm\text{std})}$ & ELO$^{(\text{CI})}$ & MASE$_{(\pm\text{std})}$ \\
\midrule
chronos-2 & 439 & $\mathbf{1339^{+60}_{-70}}$ & $0.978_{{\pm 0.56}}$ & $1297^{+70}_{-90}$ & $0.894_{{\pm 0.60}}$ \\
moirai-2-small & 413 & $1297^{+63}_{-66}$ & $1.000_{{\pm 0.46}}$ & $1290^{+83}_{-82}$ & $0.903_{{\pm 0.55}}$ \\
timesfm-2.5-200m & 471 & $1294^{+79}_{-88}$ & $1.008_{{\pm 0.81}}$ & $1247^{+79}_{-89}$ & $0.939_{{\pm 0.64}}$ \\
tirex & 580 & $1280^{+67}_{-74}$ & $0.989_{{\pm 0.67}}$ & $\mathbf{1299^{+77}_{-92}}$ & $\mathbf{0.890}_{{\pm 0.54}}$ \\
chronos-bolt-base & 728 & $1253^{+74}_{-80}$ & $1.008_{{\pm 0.73}}$ & $1131^{+96}_{-112}$ & $1.019_{{\pm 1.06}}$ \\
flowstate & 510 & $1245^{+86}_{-95}$ & $\mathbf{0.961}_{{\pm 0.52}}$ & $1203^{+91}_{-86}$ & $0.919_{{\pm 0.70}}$ \\
chronos-bolt-small & 728 & $1215^{+77}_{-90}$ & $1.022_{{\pm 0.80}}$ & $1158^{+96}_{-91}$ & $0.987_{{\pm 0.99}}$ \\
chronos-bolt-mini & 728 & $1175^{+80}_{-87}$ & $1.042_{{\pm 0.82}}$ & $1164^{+91}_{-90}$ & $0.980_{{\pm 1.06}}$ \\
moirai-base-model & 728 & $1173^{+90}_{-89}$ & $1.050_{{\pm 0.80}}$ & $1184^{+97}_{-92}$ & $0.942_{{\pm 0.54}}$ \\
chronos-bolt-tiny & 728 & $1164^{+80}_{-95}$ & $1.041_{{\pm 0.83}}$ & $1190^{+93}_{-89}$ & $0.961_{{\pm 0.91}}$ \\
sundial-128m & 696 & $1163^{+82}_{-79}$ & $1.114_{{\pm 1.18}}$ & $1120^{+91}_{-88}$ & $0.972_{{\pm 0.76}}$ \\
timesfm-2.0-500m & 728 & $1154^{+85}_{-88}$ & $1.162_{{\pm 2.48}}$ & $1187^{+86}_{-108}$ & $0.913_{{\pm 0.59}}$ \\
moirai-large & 728 & $1123^{+104}_{-93}$ & $1.147_{{\pm 1.45}}$ & $1149^{+98}_{-94}$ & $0.973_{{\pm 0.65}}$ \\
moirai-small & 728 & $1114^{+71}_{-85}$ & $1.091_{{\pm 1.06}}$ & $1107^{+94}_{-90}$ & $0.956_{{\pm 0.61}}$ \\
tinytimemixer-r1-1024-96 & 728 & $1096^{+96}_{-91}$ & $1.117_{{\pm 2.02}}$ & $1022^{+105}_{-86}$ & $1.037_{{\pm 0.94}}$ \\
tinytimemixer-r2-1024-96 & 728 & $1032^{+87}_{-79}$ & $1.230_{{\pm 2.39}}$ & $980^{+93}_{-69}$ & $1.066_{{\pm 1.20}}$ \\
visiontspp-base & 511 & $997^{+92}_{-66}$ & $1.369_{{\pm 1.12}}$ & $935^{+88}_{-87}$ & $1.162_{{\pm 1.09}}$ \\
time-moe-50m & 728 & $958^{+92}_{-77}$ & $1.455_{{\pm 5.63}}$ & $972^{+86}_{-80}$ & $1.059_{{\pm 1.18}}$ \\
time-moe-200m & 728 & $932^{+97}_{-77}$ & $1.453_{{\pm 5.25}}$ & $998^{+110}_{-90}$ & $1.068_{{\pm 1.32}}$ \\
visiontspp-large & 511 & $807^{+89}_{-68}$ & $1.744_{{\pm 2.85}}$ & $805^{+92}_{-73}$ & $1.253_{{\pm 1.42}}$ \\
seasonal-naive & 728 & $796^{+110}_{-98}$ & $1.821_{{\pm 3.13}}$ & $960^{+130}_{-136}$ & $1.184_{{\pm 0.96}}$ \\
seasonal-average & 728 & $690^{+108}_{-96}$ & $2.319_{{\pm 8.89}}$ & $902^{+115}_{-114}$ & $1.281_{{\pm 1.56}}$ \\
simple-moving-average & 728 & $531^{+75}_{-53}$ & $2.962_{{\pm 20.47}}$ & $546^{+85}_{-61}$ & $1.632_{{\pm 2.45}}$ \\
\bottomrule
\end{tabular}
\caption{ELO Leaderboard: FINGRID (15min/24h and 1h/72h)}
\label{tab:elo_fingrid_both}
\end{table*}

\section{Results Snapshot 2025-12-31}
\label{appendix:results}
This section provides a detailed breakdown of the benchmark results as of December 31, 2025. While the primary objective of TS-Arena is to facilitate a continuous and evolving evaluation of TSFMs, we present this snapshot to offer a granular view of model performance across the initial set of live data streams.
It is important to emphasize that the results presented here are intrinsically temporal and represent only a specific point in time within the live evaluation process.

A granular analysis of e.g. the SMARD live time series (Table \ref{tab:elo_smard_15min}) reveals task-dependent architectural strengths. \texttt{tirex} achieves the highest ratings in both Day-Ahead Market ($1283$) and Power Generation ($1354$) tasks, while \texttt{chronos-bolt-base} dominates Electricity Load forecasting ($1499$). The domain-specific error magnitudes indicate varying degrees of task difficulty: Power Generation forecasts remain challenging, with top models like \texttt{tirex} recording a MASE of $0.852$. In contrast, electricity load is substantially more predictable, with models such as \texttt{flowstate} achieving a MASE of $0.229$, reducing the error by over 60\% relative to the \texttt{seasonal-naive} baseline ($0.643$).

The distinct performance drop in power generation tasks highlights the limitations of applying Foundation Models as ``out-of-the-box'' universal forecasters. Unlike consumption load, generation time series frequently exhibit physical boundary conditions, such as constant zero values for solar production at night or extended periods of zero wind output. The high MASE scores in these categories suggest that current architectures, when relying solely on univariate historical contexts, struggle to model such intermittent stationarity without auxiliary data. While the integration of covariates is a planned extension of this platform, the current results imply that TSFMs cannot yet be indiscriminately applied to physically constrained time series without accounting for domain-specific characteristics.

We encourage the community to visit the TS-Arena platform for the most recent leaderboard and interactive visualizations.

\section{Backtesting Archive}
\label{appendix:archive}
To support transparency, reproducibility, and fast offline evaluation of new models, we publicly release the full TS-Arena archive on HuggingFace at \url{https://huggingface.co/datasets/DAG-UPB/TS-Arena-Archive}, with accompanying documentation at \url{https://ts-arena.live/backtesting-archive}. Each snapshot contains, for every challenge round, the historical context provided to the models, the realized ground-truth values for the forecast horizon, the forecasts submitted by every reference model.

The archive is refreshed each quarter. The current release covers data of Q1 2026. This cadence allows researchers to always evaluate against the most recent snapshot, including the newest TSFMs, without having to wait for a new benchmark publication or long-term participation.

To preserve the leakage-free nature of the evaluation in offline use, we ask researchers to follow these guidelines:
\begin{itemize}
    \item Ensure a clean temporal split in the training data: any model being evaluated against a given subset of snapshot rounds must have been trained exclusively on data observed before the earliest round in that subset. This is the offline analogue of the FPRP and prevents both direct contamination and indirect temporal leakage from correlated time series.
    \item Always cite both this paper and the specific HuggingFace snapshot version used in the evaluation.
    \item The live ts-arena.live leaderboard remains the authoritative, continuously updated ranking. The quarterly archive serves as its immutable and citable mirror for offline experiments.
\end{itemize}

\end{document}